\documentclass[10pt,twocolumn]{IEEEtran}

\usepackage{cite}
\usepackage{graphicx}
\usepackage{epsfig}
\usepackage{psfrag}
\usepackage{subfigure}
\usepackage{url}
\usepackage{stfloats}
\usepackage{amsmath}
\usepackage{color}
\usepackage{amssymb}
\usepackage{epsfig}

\usepackage{array}

\newtheorem{theorem}{Theorem}
\newtheorem{lemma}{Lemma}

\begin{document}
%
\title{Sparse Signal Recovery with Temporally Correlated Source Vectors Using Sparse Bayesian Learning}

\author{Zhilin Zhang, \IEEEmembership{Student Member, IEEE} and Bhaskar D. Rao, \IEEEmembership{Fellow, IEEE}
\thanks{Z.Zhang and B.D.Rao are with the Department of Electrical and Computer Engineering, University of California at San Diego, La Jolla, CA 92093-0407, USA. Email:\{z4zhang,brao\}@ucsd.edu. The work was supported by NSF grant CCF-0830612.} }

\markboth{To Appear in IEEE Journal of Selected Topics in Signal Processing, vol.5, no.5, 2011}{Zhang \MakeLowercase{\textit{et al.}}: }

\maketitle

\begin{abstract}
We address the sparse signal recovery problem in the context of multiple measurement vectors (MMV) when elements in each nonzero row of the solution matrix are temporally correlated. Existing algorithms do not consider such temporal correlation and thus their performance degrades significantly with the correlation. In this work, we propose a block sparse Bayesian learning framework which models the temporal correlation. We derive two sparse Bayesian learning (SBL) algorithms, which have superior recovery performance compared to existing algorithms, especially in the presence of high temporal correlation. Furthermore, our algorithms are better at handling highly underdetermined problems and require less row-sparsity on the solution matrix. We also provide analysis of the global and local minima of their cost function, and show that the SBL cost function has the very desirable property that the global minimum is at the sparsest solution to the MMV problem. Extensive experiments also provide some interesting results that motivate future theoretical research on the MMV model.
\end{abstract}


%

\IEEEpeerreviewmaketitle

\section{Introduction}
\label{sec:intro}

Sparse signal recovery, or compressed sensing, is an emerging field in signal processing \cite{Donoho2006,Candes2006,Baraniuk2007,EladBook}. The basic mathematical model is
\begin{eqnarray}
\mathbf{y}= \mathbf{\Phi} \mathbf{x} + \mathbf{v},
\label{equ:SMV model}
\end{eqnarray}
where $\mathbf{\Phi} \in \mathbb{R}^{N \times M} (N \ll M)$ is a known dictionary matrix, and any $N$ columns of $\mathbf{\Phi}$ are linearly independent (i.e. satisfies the Unique Representation Property (URP) condition \cite{Gorodnitsky1997}), $\mathbf{y} \in \mathbb{R}^{N \times 1}$ is an available measurement vector,  and $\mathbf{v}$ is an unknown  noise vector. The task is to estimate the source vector $\mathbf{x}$. To ensure a unique global solution, the number of nonzero entries in $\mathbf{x}$ has to be less than a threshold \cite{Donoho2003,Gorodnitsky1997}. This single measurement vector (SMV) model (\ref{equ:SMV model}) has a wide range of applications, such as electroencephalography (EEG)/Magnetoencephalography (MEG) source localization \cite{Gorodnitsky1995}, direction-of-arrival (DOA) estimation \cite{Malioutov2005}, radar detection \cite{Ender2010}, and magnetic resonance imaging (MRI) \cite{Gamper2008}.

Motivated by many applications such as EEG/MEG source localization and DOA estimation, where a sequence of measurement vectors are available, the basic model (\ref{equ:SMV model}) has been extended to the  multiple measurement vector (MMV) model in \cite{Rao1998,Cotter2005}, given by
\begin{eqnarray}
\mathbf{Y}= \mathbf{\Phi} \mathbf{X} + \mathbf{V},
\label{equ:MMV basicmodel}
\end{eqnarray}
where $\mathbf{Y}\triangleq[\mathbf{Y}_{\cdot1},\cdots,\mathbf{Y}_{\cdot L}] \in \mathbb{R}^{N \times L}$ is an available measurement matrix consisting of $L$ measurement vectors, $\mathbf{X}\triangleq [\mathbf{X}_{\cdot 1},\cdots,\mathbf{X}_{\cdot L}] \in \mathbb{R}^{M \times L}$ is an unknown source matrix (or called a solution matrix) with each row representing a possible source \footnote{Here for convenience we call each row in $\mathbf{X}$  a source. The term is often used in application-oriented literature. Throughout the work, the $i$-th source is denoted by $\mathbf{X}_{i\cdot}$. }, and $\mathbf{V}$ is an unknown noise matrix. A key assumption in the MMV model is that the support (i.e. indexes of nonzero entries) of every column in $\mathbf{X}$ is identical (referred as \emph{the common sparsity assumption} in literature \cite{Cotter2005}). In addition, similar to the constraint in the SMV model, the number of nonzero rows in $\mathbf{X}$ has to be below a threshold to ensure a unique and global solution \cite{Cotter2005}. This leads to the fact that $\mathbf{X}$ has a small number of nonzero rows.

It has been shown that compared to the SMV case, the successful recovery rate can be greatly improved using multiple measurement vectors \cite{Cotter2005,Eldar2009,Eldar2010AverageCase,Yuzhe:icassp}. For example, Cotter and Rao \cite{Cotter2005}  showed that by taking advantage of the MMV formulation,  one can relax the upper bound in the uniqueness condition for the solution. Tang, Eldar and their colleagues \cite{Tang2010,Eldar2010AverageCase} showed that under certain mild assumptions the recovery rate increases exponentially with the number of measurement vectors $L$. Jin and Rao \cite{Yuzhe:icassp,Yuzhe:Asilomar} analyzed the benefits of increasing $L$ by relating the MMV model to the capacity regions of MIMO communication channels. All these theoretical results reveal the advantages of the MMV model and support increasing $L$ for better recovery performance.

However, under the common sparsity assumption we cannot obtain  many measurement vectors in practical applications. The main reason is that the sparsity profile of practical signals is (slowly) time-varying, so the common sparsity assumption is valid for only a small $L$ in the MMV model. For example, in EEG/MEG source localization there is considerable evidence \cite{ElectricalNeuroimaging}  that a given pattern of dipole-source distributions \footnote{In this application the set of indexes of nonzero rows in $\mathbf{X}$ is called a pattern of dipole-source distribution.} may only exist for 10-20 ms. Since the EEG/MEG sampling frequency is generally 250 Hz, a dipole-source pattern may only exist through 5 snapshots (i.e. in the MMV model $L=5$). In DOA estimation \cite{Cotter2007},  directions of targets \footnote{In this application the index of a nonzero row in  $\mathbf{X}$ indicates a direction.} are continuously changing, and thus the source vectors that satisfy the common sparsity assumption are few. Of course, one can increase the measurement vector number at the cost of increasing the source number, but a larger source number can result in degraded recovery performance.

Thanks to  numerous algorithms for the basic SMV model, most MMV algorithms \footnote{For convenience,  algorithms for the MMV model are called MMV algorithms; algorithms for the SMV model are called SMV algorithms.} are obtained by straightforward extension of the SMV algorithms; for example, calculating the $\ell_2$ norm of each row of $\mathbf{X}$, forming a vector, and then imposing the sparsity constraint on the vector. These algorithms can be roughly divided into greedy algorithms \cite{Tropp2006SP1,Lee2011}, algorithms based on mixed norm optimization \cite{Negahban2009,Tropp2006SP2,Bach2008}, iterative reweighted algorithms \cite{David2010reweighting,Cotter2005}, and Bayesian algorithms \cite{David2007IEEE,David2010latent}.

Among the MMV algorithms, Bayesian algorithms have received much attention recently since they generally achieve the best recovery performance. Sparse Bayesian learning (SBL) is one important family of Bayesian algorithms. It was first proposed by Tipping \cite{Tipping2001,Tipping2002}, and then was greatly enriched and extended by many researchers \cite{David2004IEEE,David2007IEEE,David2010latent,David2010reweighting,Zhilin2010,Zhilin_ICASSP2011,Zhilin_ICML2011,Qiu2010,Ji2008,Tzagkarakis2010}. For example, Wipf and Rao first introduced SBL to sparse signal recovery \cite{David2004IEEE} for the SMV model, and later extended it to the MMV model, deriving the MSBL algorithm \cite{David2007IEEE}. One attraction of SBL/MSBL is that, different from the popular $\ell_1$ minimization based algorithms \cite{Lasso,BP}, whose global minimum is generally not the sparsest solution \cite{David2004IEEE,Candes2008reweighting}, the global minima of SBL/MSBL are always the sparsest one. In addition, SBL/MSBL have much fewer local minima than some classic algorithms, such as the FOCUSS family \cite{Gorodnitsky1997,Cotter2005}.

Motivated by applications where signals and other types of data often contain some kind of structures, many algorithms have been proposed \cite{Baraniuk2010,groupLasso,Zhao2009,Eldar2009}, which exploit special structures in the source matrix $\mathbf{X}$. However, most of these works focus on exploiting spatial structures (i.e. the dependency relationship among different sources) and  completely ignore   temporal structures. Besides, for tractability purposes, almost all the existing MMV algorithms (and theoretical analysis) assume that the sources are independent and identically distributed (i.i.d.) processes. This contradicts the real-world scenarios, since a practical source often has rich temporal structures. For example, the waveform smoothness  of biomedical signals has been exploited in signal processing for several decades. Besides, due to high sampling frequency, amplitudes of successive samplings of a source are strongly correlated. Recently, Zdunek and Cichocki \cite{Cichocki2008} proposed the SOB-MFOCUSS algorithm, which exploits the waveform smoothness via a pre-defined smoothness matrix. However,  the design of the smoothness matrix is completely subjective and not data-adaptive. In fact, in the task of sparse signal recovery, learning temporal structures of a source is a difficult problem. Generally, such structures are learned via a training dataset (which often contains sufficient data without noise for robust statistical inference) \cite{Cho2009,Hyvarinen2008}. Although effective for some specific signals, this method is limited. Having noticed that the temporal structures strongly affect the performance of existing algorithms, in \cite{Zhilin2010} we derived the AR-SBL algorithm, which models each source as a first-order autoregressive (AR) process and learns AR coefficients from the data per se. Although the algorithm has superior performance compared to MMV algorithms in the presence of temporal correlation, it is slow, which limits its applications. As such, there is a need for efficient algorithms that can deal more effectively with temporal correlation.

In this work, we present a block sparse Bayesian learning (bSBL) framework, which transforms the MMV model (\ref{equ:MMV basicmodel}) to a SMV model. This framework allows us to easily model the temporal correlation of sources. Based on it, we derive an algorithm, called T-SBL, which is very effective but is  slow due to its operation in a higher dimensional parameter space resulting from the MMV-to-SMV transformation. Thus, we make some approximations and derive a fast version, called T-MSBL, which operates in the original parameter space. Similar to T-SBL, T-MSBL is also effective but has much lower computational complexity. Interestingly, when compared to MSBL, the only change of T-MSBL is the replacement of $\|\mathbf{X}_{i\cdot}\|_2^2$ with the Mahalanobis distance measure, i.e. $\mathbf{X}_{i\cdot} \mathbf{B}^{-1} \mathbf{X}_{i\cdot}^T$, where $\mathbf{B}$ is a positive definite matrix estimated from data and can be partially interpreted as a covariance matrix. We analyze the global minimum and the local minima of the two algorithms' cost function. One of the key results is that in the noiseless case the global  minimum is at the sparsest solution. Extensive experiments not only show the superiority of the proposed algorithms, but also provide some interesting (even counter-intuitive) phenomena that may motivate future theoretical study.

The rest of the work is organized as follows. In Section \ref{sec:ProblemStatement} we present the bSBL  framework. In Section \ref{sec:estimateHyper} we derive the T-SBL algorithm. Its fast version, the T-MSBL algorithm, is derived in  Section \ref{sec:fastAlg}. Section \ref{sec:analysis} provides  theoretical analysis on the algorithms. Experimental results are presented in Section \ref{sec:experiment}. Finally, discussions and conclusions are drawn in the last two sections.

We introduce the notations used in this paper:
\begin{itemize}
  \item $\| \mathbf{x} \|_1,\| \mathbf{x} \|_2,\| \mathbf{A} \|_\mathcal{F}$ denote  the $\ell_1$ norm of the vector $\mathbf{x}$, the $\ell_2$ norm of $\mathbf{x}$, and the Frobenius norm of the matrix $\mathbf{A}$, respectively. $\| \mathbf{A}\|_0$ and $\| \mathbf{x} \|_0$ denote the number of nonzero rows in the matrix $\mathbf{A}$ and the number of nonzero elements in the vector $\mathbf{x}$, respectively;
  \item Bold symbols are reserved for vectors and matrices. Particularly, $\mathbf{I}_L$ denotes the identity matrix with size $L \times L$. When the dimension is evident from the context, for simplicity, we just use $\mathbf{I}$;
  \item $\mathrm{diag}\{a_1,\cdots,a_M \}$ denotes a diagonal matrix with principal diagonal elements being $a_1,\cdots,a_M$ in turn; if $\mathbf{A}_1,\cdots,\mathbf{A}_M$ are  square matrices, then $\mathrm{diag}\{\mathbf{A}_1,\cdots,\mathbf{A}_M \}$ denotes a block diagonal matrix with principal diagonal blocks being $\mathbf{A}_1,\cdots,\mathbf{A}_M$ in turn;
  \item For a matrix $\mathbf{A}$, $\mathbf{A}_{i\cdot}$ denotes the $i$-th row, $\mathbf{A}_{\cdot i}$ denotes the $i$-th column, and $\mathbf{A}_{i,j}$ denotes the element that lies in the $i$-th row and the $j$-th column;
  \item $\mathbf{A} \otimes \mathbf{B}$ represents the Kronecker product of the two matrices $\mathbf{A}$ and $\mathbf{B}$. $\mathrm{vec}(\mathbf{A})$ denotes the vectorization of the matrix $\mathbf{A}$ formed by stacking its columns into a single column vector. $\mathrm{Tr}(\mathbf{A})$ denotes the trace of $\mathbf{A}$. $\mathbf{A}^T$ denotes the transpose of $\mathbf{A}$.
\end{itemize}

\section{Block Sparse Bayesian Learning Framework}
\label{sec:ProblemStatement}

Most  existing works do not deal with the temporal correlation of sources. For many non-Bayesian algorithms, incorporating temporal correlation is not easy due to the lack of a well defined methodology to modify the diversity measures employed in the optimization procedure. For example, it is not clear how to best incorporate correlation in $\ell_1$ norm based methods. For this reason, we adopt a  probabilistic approach to incorporate correlation structure. Particularly, we have found it convenient to incorporate correlation into the  sparse Bayesian learning (SBL) methodology.

Initially, SBL was proposed for regression and classification in machine learning \cite{Tipping2001}. Then Wipf and Rao \cite{David2004IEEE} applied it to the SMV model (\ref{equ:SMV model}) for sparse signal recovery. The idea is to find the posterior probability $p(\mathbf{x}|\mathbf{y};\Theta)$ via the Bayesian rule, where $\Theta$ indicates the set of all the hyperparameters. Given the hyperparameters, the solution $\widehat{\mathbf{x}}$ is given by the Maximum-A-Posterior (MAP) estimate. The hyperparameters are estimated from data by marginalizing over $\mathbf{x}$ and then performing evidence maximization or Type-II Maximum Likelihood \cite{Tipping2001}. To solve the MMV problem (\ref{equ:MMV basicmodel}), Wipf and Rao \cite{David2007IEEE} proposed the MSBL algorithm, which implicitly applies the $\ell_2$ norm on each source $\mathbf{X}_{i\cdot}$. One drawback of this algorithm is that the temporal correlation of sources is not exploited to improve performance.

To exploit the temporal correlation, we propose another SBL framework, called the block sparse Bayesian learning (bSBL) framework. In this framework, the MMV model is transformed to a block SMV model. In this way, we can easily model the temporal correlation of sources and derive new algorithms.

First, we assume all the sources $\mathbf{X}_{i\cdot}$ ($\forall i$) are mutually independent, and the density of each $\mathbf{X}_{i\cdot}$ is Gaussian, given by
\begin{eqnarray}
p(\mathbf{X}_{i\cdot}; \gamma_i,\mathbf{B}_i) \sim \mathcal{N}(\mathbf{0},\gamma_i \mathbf{B}_i), \quad  i=1,\cdots,M
\nonumber
\end{eqnarray}
where $\gamma_i$ is a nonnegative hyperparameter controlling the row sparsity of $\mathbf{X}$ as in the basic SBL \cite{Tipping2001,David2004IEEE,David2007IEEE}. When $\gamma_i = 0$,  the associated $\mathbf{X}_{i\cdot}$ becomes zeros. $\mathbf{B}_i$ is a positive definite  matrix that captures the  correlation structure of $\mathbf{X}_{i\cdot}$ and needs to be estimated.

By letting $\mathbf{y}=\mathrm{vec}(\mathbf{Y}^T) \in \mathbb{R}^{NL \times 1}$, $\mathbf{D} = \mathbf{\Phi} \otimes \mathbf{I}_L$, $\mathbf{x}=\mathrm{vec}(\mathbf{X}^T) \in \mathbb{R}^{ML \times 1}$, $\mathbf{v}=\mathrm{vec}(\mathbf{V}^T)$, we can transform the MMV model   to the block SMV model
\begin{eqnarray}
\mathbf{y}= \mathbf{D} \mathbf{x} + \mathbf{v}.
\label{equ:blocksparsemodel}
\end{eqnarray}
To elaborate the block sparsity model (\ref{equ:blocksparsemodel}), we rewrite it as $\mathbf{y}= [\boldsymbol{\phi}_1 \otimes \mathbf{I}_L, \cdots, \boldsymbol{\phi}_M \otimes \mathbf{I}_L] [\mathbf{x}_1^T,\cdots,\mathbf{x}_M^T]^T + \mathbf{v} = \sum_{i=1}^M (\boldsymbol{\phi}_i \otimes \mathbf{I}_L) \mathbf{x}_i + \mathbf{v}$, where $\boldsymbol{\phi}_i$ is the $i$-th column in $\mathbf{\Phi}$, and $\mathbf{x}_i \in \mathbb{R}^{L \times 1}$ is the $i$-th block in $\mathbf{x}$ and $\mathbf{x}_i= \mathbf{X}_{i\cdot}^T$.  $K$ nonzero rows in $\mathbf{X}$ means $K$ nonzero blocks in $\mathbf{x}$. Thus $\mathbf{x}$ is block-sparse.

Assume elements in the noise vector $\mathbf{v}$ are independent and each has a Gaussian distribution, i.e. $p(v_i) \sim \mathcal{N}(0,\lambda)$, where $v_i$ is the $i$-th element in $\mathbf{v}$ and $\lambda$ is the variance. For the block model (\ref{equ:blocksparsemodel}), the Gaussian likelihood  is
\begin{eqnarray}
p(\mathbf{y}|\mathbf{x}; \lambda) \sim \mathcal{N}_{y|x}(\mathbf{D}\mathbf{x}, \lambda \mathbf{I}).
\nonumber
\end{eqnarray}
The prior for $\mathbf{x}$ is given by
\begin{eqnarray}
p(\mathbf{x}; \gamma_i,\mathbf{B}_i,\forall i)
 \sim  \mathcal{N}_x(\textbf{0},\mathbf{\Sigma}_0),
\nonumber
\end{eqnarray}
where  $\mathbf{\Sigma}_0$ is
\begin{equation}
\mathbf{\Sigma}_0 =  \left[
\begin {array}{cccc}
\gamma_1\mathbf{B}_1  &                        & \\
         &            \ddots      & \\
         &                        & \gamma_M \mathbf{B}_M
\end {array}
\right].
\label{equ:Sigma0}
\end{equation}
Using the Bayes rule we obtain the posterior density of $\mathbf{x}$, which is also Gaussian,
\begin{eqnarray}
p(\mathbf{x}|\mathbf{y}; \lambda, \gamma_i,\mathbf{B}_i,\forall i) = \mathcal{N}_x (\boldsymbol{\mu}_x, \mathbf{\Sigma}_x)
\nonumber
\end{eqnarray}
with the mean
\begin{eqnarray}
\boldsymbol{\mu}_x = \frac{1}{\lambda} \mathbf{\Sigma}_x \mathbf{D}^T \mathbf{y}
\label{equ:mu_x}
\end{eqnarray}
and the covariance matrix
\begin{eqnarray}
\mathbf{\Sigma}_x &=& (\mathbf{\Sigma}_0^{-1} + \frac{1}{\lambda}\mathbf{D}^T \mathbf{D})^{-1} \nonumber \\
&=& \mathbf{\Sigma}_0 - \mathbf{\Sigma}_0 \mathbf{D}^T \big(\lambda \mathbf{I} + \mathbf{D} \mathbf{\Sigma}_0 \mathbf{D}^T \big)^{-1} \mathbf{D} \mathbf{\Sigma}_0.
\label{equ:Sigma_x}
\end{eqnarray}

So given all the hyperparameters $\lambda, \gamma_i,\mathbf{B}_i,\forall i$, the MAP estimate of $\mathbf{x}$ is given by:
\begin{eqnarray}
\mathbf{x}^* \triangleq \boldsymbol{\mu}_x &=& (\lambda \mathbf{\Sigma}_0^{-1} + \mathbf{D}^T \mathbf{D})^{-1} \mathbf{D}^T \mathbf{y} \nonumber \\
&=& \mathbf{\Sigma}_0 \mathbf{D}^T \big(\lambda \mathbf{I} +  \mathbf{D} \mathbf{\Sigma}_0 \mathbf{D}^T \big)^{-1} \mathbf{y}
\label{equ:noisySolution}
\end{eqnarray}
where the last equation follows the matrix identity $(\mathbf{I} + \mathbf{AB})^{-1} \mathbf{A} \equiv \mathbf{A}(\mathbf{I}+\mathbf{BA})^{-1}$, and $\mathbf{\Sigma}_0$ is the block diagonal matrix given by (\ref{equ:Sigma0}) with many diagonal block matrices being zeros. Clearly, the block sparsity of $\mathbf{x}^*$ is  controlled by the $\gamma_i$'s in $\mathbf{\Sigma}_0$: during the learning procedure, when $\gamma_k = 0$, the associated $k$-th block in $\mathbf{x}^*$ becomes zeros, and the associated dictionary vectors $\boldsymbol{\phi}_k \otimes \mathbf{I}_L$ are pruned out \footnote{In practice, we judge whether $\gamma_k$ is less than a small threshold, e.g. $10^{-5}$. If it is, then the associated dictionary vectors are pruned out from the learning procedure and the associated block in $\mathbf{x}$ is set to zeros.}.

To estimate the hyperparameters we can use evidence maximization or Type-II maximum likelihood \cite{Tipping2001}. This involves marginalizing over the weights $\mathbf{x}$ and then performing maximum likelihood estimation.  We refer to the whole framework including the solution (\ref{equ:noisySolution})  and the hyperparameter estimation as the  block sparse Bayesian learning (bSBL) framework. Note that in contrast to the original SBL framework, the bSBL framework models the temporal structures of sources in the prior density via the matrices $\mathbf{B}_i$ ($i=1,\cdots,M$). Different ways to learn the matrices result in different algorithms. We will discuss the learning of these matrices and other hyperparameters in the following sections.

\section{Estimation of Hyperparameters}
\label{sec:estimateHyper}

Before estimating the hyperparameters, we note that assigning a  different  matrix $\mathbf{B}_i$ to each source $\mathbf{X}_{i\cdot}$ will result in overfitting \cite{Cawley2007,Guyon2010}  due to limited data and too many parameters. To avoid the overfitting, we  consider using  one positive definite matrix $\mathbf{B}$ to model all the source covariance matrices up to a scalar \footnote{Note that the covariance matrix in the density of $\mathbf{X}_{i\cdot}$ is $\gamma_i \mathbf{B}_i$. }. Thus Eq.(\ref{equ:Sigma0}) becomes $\mathbf{\Sigma}_0 = \mathbf{\Gamma} \otimes \mathbf{B}$ with $\mathbf{\Gamma} \triangleq \mathrm{diag}(\gamma_1,\cdots,\gamma_M)$. Although this strategy is equivalent to assuming all the sources have the same correlation structure, it leads to very good results even if all the sources have totally different correlation structures (see Section \ref{sec:experiment}). More importantly, this constraint does not destroy the global minimum property (i.e. the global unique solution is the sparsest solution) of our algorithms, as confirmed by Theorem \ref{theorem:global} in Section \ref{sec:analysis}.

To find the hyperparameters $\Theta =\{  \gamma_1,\cdots,\gamma_M, \mathbf{B}, \lambda\}$,  we employ the Expectation-Maximization (EM) method to maximize $p(\mathbf{y};\Theta)$. This is equivalent to minimizing $-\log p(\mathbf{y};\Theta)$, yielding the effective cost function:
\begin{eqnarray}
\mathcal{L}(\Theta) = \mathbf{y}^T \mathbf{\Sigma}_y^{-1} \mathbf{y} + \log|\mathbf{\Sigma}_y|,
\label{equ:costfunc}
\end{eqnarray}
where $\mathbf{\Sigma}_y \triangleq \lambda \mathbf{I}  + \mathbf{D}\mathbf{\Sigma}_0 \mathbf{D}^T$. The  EM formulation proceeds by treating  $\mathbf{x}$ as hidden variables and then maximizing:
\begin{eqnarray}
Q(\Theta) &=& E_{x|y;\Theta^{(\mathrm{old})}} \big[ \log p(\mathbf{y},\mathbf{x}; \Theta)  \big]    \nonumber \\
&=& E_{x|y;\Theta^{(\mathrm{old})}} \big[ \log p(\mathbf{y}|\mathbf{x};\lambda)  \big] \nonumber \\
&& + E_{x|y;\Theta^{(\mathrm{old})}} \big[ \log p(\mathbf{x};\gamma_1,\cdots,\gamma_M,\mathbf{B}) \big] \label{equ:Qfunction}
\end{eqnarray}
where $\Theta^{(\mathrm{old})}$ denotes the estimated hyperparameters in the previous iteration.

To estimate $\boldsymbol{\gamma} \triangleq [\gamma_1,\cdots,\gamma_M]$ and $\mathbf{B}$, we notice that the first term in (\ref{equ:Qfunction}) is unrelated to $\boldsymbol{\gamma}$ and $\mathbf{B}$. So, the Q function (\ref{equ:Qfunction}) can be simplified to:
\begin{eqnarray}
Q(\boldsymbol{\gamma},\mathbf{B}) = E_{x|y;\Theta^{(\mathrm{old})}} \big[ \log p(\mathbf{x};\boldsymbol{\gamma},\mathbf{B}) \big]. \nonumber
\end{eqnarray}
It can be shown that\footnote{The $\propto$ notation is used to indicate that  terms that do not contribute to the subsequent optimization of the
parameters have been dropped. This convention will be followed through out the paper.}
\begin{eqnarray}
\log p(\mathbf{x};\boldsymbol{\gamma},\mathbf{B})
\propto -\frac{1}{2} \log \big(|\mathbf{\Gamma}|^L|\mathbf{B}|^M \big) - \frac{1}{2} \mathbf{x}^T (\mathbf{\Gamma}^{-1} \otimes \mathbf{B}^{-1}) \mathbf{x}, \nonumber
\end{eqnarray}
which results in
\begin{eqnarray}
Q(\boldsymbol{\gamma},\mathbf{B})
&\propto & -\frac{L}{2} \log \big(|\mathbf{\Gamma}| \big) - \frac{M}{2} \log \big( |\mathbf{B}| \big) \nonumber \\
&& - \frac{1}{2} \mathrm{Tr}\Big[ \big(\mathbf{\Gamma}^{-1} \otimes \mathbf{B}^{-1} \big) \big(  \mathbf{\Sigma}_x + \boldsymbol{\mu}_x  \boldsymbol{\mu}_x^T   \big)\Big],
\label{equ:Qfunction_gamma_B}
\end{eqnarray}
where $\boldsymbol{\mu}_x$ and $\mathbf{\Sigma}_x$  are evaluated according to (\ref{equ:mu_x}) and (\ref{equ:Sigma_x}), given the estimated hyperparameters $\Theta^{(\mathrm{old})}$.

The derivative of  (\ref{equ:Qfunction_gamma_B}) with respect to $\gamma_i \;(i=1,\cdots,M)$ is given by
\begin{eqnarray}
\frac{\partial Q}{\partial \gamma_i}
= -\frac{L}{2\gamma_i} + \frac{1}{2\gamma_i^2} \mathrm{Tr}\Big[ \mathbf{B}^{-1} \big( \mathbf{\Sigma}_x^i + \boldsymbol{\mu}_x^i ( \boldsymbol{\mu}_x^i)^T   \big)\Big], \nonumber
\end{eqnarray}
where we define (using the MATLAB notations)
\begin{equation}
\left\{
\begin{array}{ll}
\boldsymbol{\mu}_x^i \triangleq \boldsymbol{\mu}_x ((i-1)L+1 \;:\; iL) \\
\mathbf{\Sigma}_x^{i} \triangleq  \mathbf{\Sigma}_x ((i-1)L+1 \: :\: iL \;,\; (i-1)L+1 \::\: iL)
\end{array}
\right.
\label{equ:meanVarianceBlockDefinition}
\end{equation}
So the learning rule for $\gamma_i \;(i=1,\cdots,M)$ is given by
\begin{eqnarray}
\gamma_i \leftarrow \frac{\mathrm{Tr}\big[ \mathbf{B}^{-1} \big( \mathbf{\Sigma}_x^i + \boldsymbol{\mu}_x^i (\boldsymbol{\mu}_x^i)^T\big)\big]}{L}, \quad i=1,\cdots,M
\label{equ:gammaRule}
\end{eqnarray}

On the other hand, the gradient of (\ref{equ:Qfunction_gamma_B}) over $\mathbf{B}$ is given by
\begin{eqnarray}
\frac{\partial Q}{\partial \mathbf{B}}
=  - \frac{M}{2} \mathbf{B}^{-1} +  \frac{1}{2} \sum_{i=1}^M \frac{1}{\gamma_i} \mathbf{B}^{-1}  \big( \mathbf{\Sigma}_x^i + \boldsymbol{\mu}_x^i (\boldsymbol{\mu}_x^i)^T\big)  \mathbf{B}^{-1}. \nonumber
\end{eqnarray}
Thus we obtain the learning rule for $\mathbf{B}$:
\begin{eqnarray}
\mathbf{B} \leftarrow \frac{1}{M} \sum_{i=1}^M \frac{\mathbf{\Sigma}_x^i + \boldsymbol{\mu}_x^i (\boldsymbol{\mu}_x^i)^T}{\gamma_i}.
\label{equ:Brule}
\end{eqnarray}

To estimate $\lambda$, the Q function (\ref{equ:Qfunction}) can be simplified to
\begin{eqnarray}
Q(\lambda) &=&  E_{x|y;\Theta^{(\mathrm{old})}} \big[ \log p(\mathbf{y}|\mathbf{x};\lambda)  \big]   \nonumber \\
&\propto & -\frac{NL}{2} \log \lambda - \frac{1}{2\lambda} E_{x|y;\Theta^{(\mathrm{old})}} \big[ \|\mathbf{y}-\mathbf{Dx}   \|_2^2\big] \nonumber \\
&= &  -\frac{NL}{2} \log \lambda -\frac{1}{2\lambda} \Big[\|\mathbf{y}-\mathbf{D}\boldsymbol{\mu}_x\|_2^2 \nonumber \\
&& + E_{x|y;\Theta^{(\mathrm{old})}}\big[\|\mathbf{D}(\mathbf{x}-\boldsymbol{\mu}_x)\|_2^2 \big]  \Big]   \nonumber \\
&=&  -\frac{NL}{2} \log \lambda -\frac{1}{2\lambda} \Big[\|\mathbf{y}-\mathbf{D}\boldsymbol{\mu}_x\|_2^2 + \mathrm{Tr}\big( \mathbf{\Sigma}_x \mathbf{D}^T \mathbf{D} \big)  \Big]  \nonumber \\
&=& -\frac{NL}{2} \log \lambda -\frac{1}{2\lambda} \Big[\|\mathbf{y}-\mathbf{D}\boldsymbol{\mu}_x\|_2^2 \nonumber \\
&& + \widehat{\lambda} \mathrm{Tr}\big( \mathbf{\Sigma}_x (\mathbf{\Sigma}_x^{-1}-\mathbf{\Sigma}_0^{-1}) \big)  \Big]  \label{equ:comment1} \\
&=& -\frac{NL}{2} \log \lambda -\frac{1}{2\lambda} \Big[\|\mathbf{y}-\mathbf{D}\boldsymbol{\mu}_x\|_2^2 \nonumber \\
&& + \widehat{\lambda} \big[ML-\mathrm{Tr}(\mathbf{\Sigma}_x \mathbf{\Sigma}_0^{-1})\big]   \Big], \label{equ:Qfunction_lambda}
\end{eqnarray}
where (\ref{equ:comment1}) follows from  the first equation in (\ref{equ:Sigma_x}), and $\widehat{\lambda}$ denotes the  estimated $\lambda$ in the previous iteration. The $\lambda$ learning rule is obtained by setting the derivative of (\ref{equ:Qfunction_lambda}) over $\lambda$ to zero, leading to
\begin{eqnarray}
\lambda  \leftarrow \frac{\|\mathbf{y}-\mathbf{D} \boldsymbol{\mu}_x\|_2^2 + \lambda \big[ML-\mathrm{Tr}( \mathbf{\Sigma}_x \mathbf{\Sigma}_0^{-1})\big]}{NL},
\label{equ:lambdaRule}
\end{eqnarray}
where the $\lambda$ on the right-hand side is the $\widehat{\lambda}$ in (\ref{equ:Qfunction_lambda}). There are some challenges to estimate $\lambda$ in SMV models. This, however, is alleviated in MMV models when considering temporal correlation. We elaborate on this next.

In the SBL framework (either for the SMV model or for the MMV model), many learning rules for $\lambda$ have been derived \cite{Tipping2001,David2004IEEE,David2007IEEE,Qiu2010}. However, in noisy environments some of the learning rules probably cannot provide an optimal $\lambda$, thus leading to degraded performance. For the basic SBL/MSBL algorithms, Wipf et al \cite{David2007IEEE} pointed out that the reason is that $\lambda$ and appropriate $N$ nonzero hyperparameters $\gamma_i$ make an identical contribution to the covariance $\mathbf{\Sigma}_y = \lambda \mathbf{I} + \mathbf{\Phi} \mathbf{\Gamma} \mathbf{\Phi}^T$ in the cost functions of SBL/MSBL. To explain this, they gave an example: let a dictionary matrix $\mathbf{\Phi}'=[\mathbf{\Phi}_0,\mathbf{I}]$, where $\mathbf{\Phi}' \in \mathbb{R}^{N \times M}$ and $\mathbf{\Phi}_0 \in \mathbb{R}^{N \times (M-N)}$. Then the $\lambda$ as well as the $N$ hyperparameters $\{\gamma_{M-N+1},\cdots,\gamma_{M}\}$ associated with the columns of the identity matrix in $\Phi'$ are not identifiable, because
\begin{eqnarray}
\mathbf{\Sigma}_y &=& \lambda \mathbf{I} + \mathbf{\Phi}' \mathbf{\Gamma} {\mathbf{\Phi}'}^T \nonumber \\
&=& \lambda \mathbf{I}  + [ \mathbf{\Phi}_0,\mathbf{I}  ] \mathrm{diag}\{\gamma_1,\cdots,\gamma_M  \} [ \mathbf{\Phi}_0,\mathbf{I}  ]^T \nonumber \\
&=& \lambda \mathbf{I} + \mathbf{\Phi}_0 \mathrm{diag}\{\gamma_1,\cdots,\gamma_{M-N}  \} \mathbf{\Phi}_0^T \nonumber \\
&& + \mathrm{diag}\{ \gamma_{M-N+1},\cdots,\gamma_{M} \} \nonumber
\end{eqnarray}
indicating a nonzero value of $\lambda$ and  appropriate values of the $N$ nonzero hyperparameters, i.e. $\gamma_{M-N+1},\cdots,\gamma_{M}$, can make an identical contribution to the covariance matrix $\mathbf{\Sigma}_y$. This problem can be worse when the noise covariance matrix is $\mathrm{diag}(\lambda_1,\cdots,\lambda_N)$ with arbitrary nonzero $\lambda_i$, instead of $\lambda \mathbf{I}$.

However, our learning rule (\ref{equ:lambdaRule}) does not have such ambiguity problem. To see this, we now examine the covariance matrix $\mathbf{\Sigma}_y $ in our cost function (\ref{equ:costfunc}). Noting that $\mathbf{D} = \mathbf{\Phi}' \otimes \mathbf{I}$, we have
\begin{eqnarray}
\mathbf{\Sigma}_y &=& \lambda \mathbf{I} + \mathbf{D} \mathbf{\Sigma}_0 \mathbf{D}^T \nonumber \\
&=& \lambda \mathbf{I} + (\mathbf{\Phi}' \otimes \mathbf{I}) \big(\mathrm{diag}\{\gamma_1,\cdots,\gamma_M  \} \otimes \mathbf{B} \big) (\mathbf{\Phi}' \otimes \mathbf{I})^T \nonumber \\
&=& \lambda \mathbf{I} +  [ \mathbf{\Phi}_0 \otimes \mathbf{I}, \mathbf{I} \otimes \mathbf{I}  ] \big(\mathrm{diag}\{\gamma_1,\cdots,\gamma_M  \} \otimes \mathbf{B}\big) \nonumber \\
 && \cdot [\mathbf{\Phi}_0 \otimes \mathbf{I}, \mathbf{I} \otimes \mathbf{I}]^T  \nonumber \\
&=& \lambda \mathbf{I} + (\mathbf{\Phi}_0 \mathrm{diag}\{\gamma_1,\cdots,\gamma_{M-N}\} \mathbf{\Phi}_0^T) \otimes \mathbf{B} \nonumber \\
&& + \mathrm{diag}\{ \gamma_{M-N+1},\cdots,\gamma_{M} \} \otimes \mathbf{B}.  \nonumber
\end{eqnarray}
Obviously, since $\mathbf{B}$ is not an identity matrix \footnote{Note that even all the sources are i.i.d. processes, the estimated $\mathbf{B}$ in practice is not an exact identity matrix.}, $\lambda$ and $\{ \gamma_{M-N+1},\cdots,\gamma_{M} \}$  cannot identically contribute to $\mathbf{\Sigma}_y$.

The SBL algorithm using the learning rules (\ref{equ:Sigma_x}), (\ref{equ:noisySolution}), (\ref{equ:gammaRule}), (\ref{equ:Brule}) and (\ref{equ:lambdaRule}) is denoted by \emph{\textbf{T-SBL}}.

\section{An Efficient Algorithm Processing in the Original Problem Space}
\label{sec:fastAlg}

The proposed T-SBL algorithm has excellent performance in terms of recovery performance (see Section \ref{sec:experiment}). But it is not fast because it learns the parameters in a higher dimensional space instead of the original problem space \footnote{T-SBL can be directly used to solve the block sparsity models \cite{Negahban2009,groupLasso,Eldar2009}. In this case, the algorithm directly performs in the original parameter space and thus it is not slow (compared to the speed of some other algorithms for the block sparsity models).}. For example, the dictionary matrix is of the size  $NL \times ML$ in the bSBL framework, while it is only of the size $N \times M$ in the original MMV model. Interestingly, the MSBL developed for i.i.d. sources has complexity $\mathcal{O}(N^2 M)$ and does not exhibit this drawback \cite{David2007IEEE}. Motivated by this, we make a reasonable approximation and  back-map  T-SBL to the original space \footnote{By back-mapping, we mean we use some approximation to simplify the algorithm such that the simplified version directly operates in the parameter space of the original MMV model.}.

For convenience, we first list the  MSBL algorithm derived in \cite{David2007IEEE}:
\begin{eqnarray}
\mathbf{\Xi}_x &= & \big( \mathbf{\Gamma}^{-1} + \frac{1}{\lambda} \mathbf{\Phi}^T \mathbf{\Phi}   \big)^{-1}  \label{equ:MSBL_cov}\\
\mathbf{X} &=& \mathbf{\Gamma} \mathbf{\Phi}^T \big( \lambda \mathbf{I} + \mathbf{\Phi} \mathbf{\Gamma} \mathbf{\Phi}^T \big)^{-1} \mathbf{Y}  \label{equ:X_est_MSBL} \\
\gamma_i &=& \frac{1}{L} \| \mathbf{X}_{i \cdot} \|_2^2 + (\mathbf{\Xi}_x)_{ii}, \quad \forall i
\label{equ:gammaRuleInMSBL}
\end{eqnarray}
An important observation is the lower dimension of the matrix operations involved in this algorithm. We  attempt to achieve similar complexity for the T-SBL algorithm by adopting the following approximation:
\begin{eqnarray}
\big(\lambda \mathbf{I}_{NL} +  \mathbf{D} \mathbf{\Sigma}_0 \mathbf{D}^T \big)^{-1} &=& \big( \lambda \mathbf{I}_{NL} + (\mathbf{\Phi} \mathbf{\Gamma} \mathbf{\Phi}^T) \otimes \mathbf{B}   \big)^{-1} \nonumber \\
&\approx & \big(\lambda \mathbf{I}_N +  \mathbf{\Phi} \mathbf{\Gamma} \mathbf{\Phi}^T \big)^{-1} \otimes \mathbf{B}^{-1} \nonumber \\
\label{equ:approximationFormula}
\end{eqnarray}
which is exact when  $\lambda = 0$ or $\mathbf{B} = \mathbf{I}_L$. For high signal-to-noise ratio (SNR) or low correlation the
approximation is quite reasonable. But our experiments will show that our algorithm adopting this approximation performs quite well over a broader range of conditions (see Section \ref{sec:experiment}).

Now we use the approximation to simplify the $\gamma_i$ learning rule (\ref{equ:gammaRule}). First, we consider the following term in (\ref{equ:gammaRule}):
\begin{eqnarray}
\frac{1}{L}\mathrm{Tr}\big( \mathbf{B}^{-1}  \mathbf{\Sigma}_x^i  \big) &=& \frac{1}{L} \mathrm{Tr}\Big[ \gamma_i \mathbf{I}_L - \gamma_i^2 (\boldsymbol{\phi}_i^T \otimes \mathbf{I}_L) (\lambda \mathbf{I}_{NL} + \nonumber \\
&&   \mathbf{D} \mathbf{\Sigma}_0 \mathbf{D}^T )^{-1} (\boldsymbol{\phi}_i \otimes \mathbf{I}_L) \cdot \mathbf{B}  \Big] \label{equ:comment66} \\
&\approx & \gamma_i - \frac{\gamma_i^2}{L} \mathrm{Tr}\Big[ \Big( \big[ \boldsymbol{\phi}_i^T \big(\lambda \mathbf{I}_N + \nonumber \\
 &&   \mathbf{\Phi} \mathbf{\Gamma} \mathbf{\Phi}^T \big)^{-1}  \boldsymbol{\phi}_i \big] \otimes \mathbf{B}^{-1} \Big) \mathbf{B} \Big]  \nonumber \\
& = & \gamma_i - \frac{\gamma_i^2}{L} \mathrm{Tr}\Big[ \Big( \boldsymbol{\phi}_i^T \big(\lambda \mathbf{I}_N + \nonumber \\
&&   \mathbf{\Phi} \mathbf{\Gamma} \mathbf{\Phi}^T \big)^{-1}  \boldsymbol{\phi}_i \Big) \mathbf{I}_L \Big] \nonumber \\
&=& \gamma_i - \gamma_i^2 \boldsymbol{\phi}_i^T \big(\lambda \mathbf{I}_N + \mathbf{\Phi} \mathbf{\Gamma} \mathbf{\Phi}^T  \big)^{-1} \boldsymbol{\phi}_i \nonumber \\
&=& (\mathbf{\Xi}_x)_{ii}
\label{equ:equivalent_1}
\end{eqnarray}
where (\ref{equ:comment66})  follows the second equation in (\ref{equ:Sigma_x}), and $\mathbf{\Xi}_x$ is given in (\ref{equ:MSBL_cov}). Using the same approximation (\ref{equ:approximationFormula}), the $\boldsymbol{\mu}_x$ in (\ref{equ:gammaRule}) can be expressed as
\begin{eqnarray}
\boldsymbol{\mu}_x &\approx& (\mathbf{\Gamma} \otimes \mathbf{B})(\mathbf{\Phi}^T \otimes \mathbf{I}) \nonumber \\
&& \cdot \big[ \big(\lambda \mathbf{I} +  \mathbf{\Phi} \mathbf{\Gamma} \mathbf{\Phi}^T \big)^{-1} \otimes \mathbf{B}^{-1} \big] \mathrm{vec}(\mathbf{Y}^T) \label{equ:comment3} \\
&=& \big[ \mathbf{\Gamma}\mathbf{\Phi}^T  \big(\lambda \mathbf{I} +  \mathbf{\Phi} \mathbf{\Gamma} \mathbf{\Phi}^T \big)^{-1}  \big] \otimes \mathbf{I} \cdot \mathrm{vec}(\mathbf{Y}^T)  \nonumber \\
&=& \mathrm{vec} \big( \mathbf{Y}^T  \big(\lambda \mathbf{I} +  \mathbf{\Phi} \mathbf{\Gamma} \mathbf{\Phi}^T \big)^{-1} \mathbf{\Phi} \mathbf{\Gamma} \big) \nonumber \\
&=& \mathrm{vec} (\mathbf{X}^T)
\label{equ:equivalent_2}
\end{eqnarray}
where (\ref{equ:comment3}) follows (\ref{equ:mu_x}) and the approximation (\ref{equ:approximationFormula}), and $\mathbf{X}$ is given in (\ref{equ:X_est_MSBL}). Therefore, based on (\ref{equ:equivalent_1}) and  (\ref{equ:equivalent_2}), we can transform the $\gamma_i$ learning rule (\ref{equ:gammaRule})  to the following form:
\begin{eqnarray}
\gamma_i \leftarrow \frac{1}{L} \mathbf{X}_{i \cdot} \mathbf{B}^{-1} \mathbf{X}_{i \cdot}^T + (\mathbf{\Xi}_x)_{ii}, \quad \forall i
\label{equ:gammaRule_tSBL}
\end{eqnarray}

To simplify the $\mathbf{B}$ learning rule (\ref{equ:Brule}), we note that
\begin{eqnarray}
\mathbf{\Sigma}_x &=& \mathbf{\Sigma}_0 - \mathbf{\Sigma}_0 \mathbf{D}^T (\lambda \mathbf{I} + \mathbf{D}\mathbf{\Sigma}_0 \mathbf{D}^T)^{-1} \mathbf{D} \mathbf{\Sigma}_0 \nonumber \\
&=& \mathbf{\Gamma} \otimes \mathbf{B} - (\mathbf{\Gamma} \otimes \mathbf{B}) (\mathbf{\Phi}^T \otimes \mathbf{I}) (\lambda \mathbf{I} + \mathbf{D}\mathbf{\Sigma}_0 \mathbf{D}^T)^{-1} \nonumber \\
&& \cdot (\mathbf{\Phi} \otimes \mathbf{I}) (\mathbf{\Gamma} \otimes \mathbf{B}) \nonumber \\
&\thickapprox & \mathbf{\Gamma} \otimes \mathbf{B} - \big[(\mathbf{\Gamma} \mathbf{\Phi}^T)\otimes \mathbf{B} \big] \big[ ( \lambda \mathbf{I} + \mathbf{\Phi} \mathbf{\Gamma} \mathbf{\Phi}^T )^{-1} \otimes \mathbf{B}^{-1}\big] \nonumber \\
&& \cdot \big[(\mathbf{\Phi}\mathbf{\Gamma}) \otimes \mathbf{B} \big]  \label{equ:comment5} \\
&=& \big( \mathbf{\Gamma} - \mathbf{\Gamma}\mathbf{\Phi}^T( \lambda \mathbf{I} + \mathbf{\Phi} \mathbf{\Gamma} \mathbf{\Phi}^T )^{-1} \mathbf{\Phi}\mathbf{\Gamma}   \big) \otimes \mathbf{B} \nonumber \\
&=& \mathbf{\Xi}_x \otimes \mathbf{B}, \nonumber
\end{eqnarray}
where (\ref{equ:comment5}) uses the approximation (\ref{equ:approximationFormula}). Using the definition (\ref{equ:meanVarianceBlockDefinition}), we have $\mathbf{\Sigma}_x^i = (\mathbf{\Xi}_x)_{ii} \mathbf{B}$. Therefore, the learning rule (\ref{equ:Brule}) becomes:
\begin{eqnarray}
\mathbf{B} &\leftarrow& \Big( \frac{1}{M}\sum_{i=1}^M \frac{(\mathbf{\Xi}_x)_{ii}}{\gamma_i}  \Big) \mathbf{B} + \frac{1}{M}\sum_{i=1}^M \frac{\mathbf{X}_{i \cdot}^T \mathbf{X}_{i \cdot} }{\gamma_i}.
\end{eqnarray}
From the learning rule above, we can directly construct a fixed-point learning rule, given by
\begin{eqnarray}
\mathbf{B}  &\leftarrow& \frac{1}{M (1-\rho)}\sum_{i=1}^M \frac{\mathbf{X}_{i \cdot}^T \mathbf{X}_{i \cdot} }{\gamma_i} \nonumber
\end{eqnarray}
where $\rho = \frac{1}{M}\sum_{i=1}^M \gamma_i^{-1} (\mathbf{\Xi}_x)_{ii}$. To increase the robustness, however, we suggest using the rule below:
\begin{eqnarray}
\widetilde{\mathbf{B}} &\leftarrow&  \sum_{i=1}^M \frac{\mathbf{X}_{i \cdot}^T \mathbf{X}_{i \cdot} }{\gamma_i}  \label{equ:Brule_simplified}\\
\mathbf{B} &\leftarrow& \widetilde{\mathbf{B}}/ \| \widetilde{\mathbf{B}}  \|_\mathcal{F} \label{equ:Brule_norm}
\end{eqnarray}
where (\ref{equ:Brule_norm}) is to remove the ambiguity between $\mathbf{B}$ and $\gamma_i$ ($\forall i$). This learning rule performs well in high SNR cases and noiseless cases \footnote{Note that in (\ref{equ:Brule_simplified}) when the number of distinct nonzero rows in $\mathbf{X}$ is smaller than the number of measurement vectors, the matrix $\widetilde{\mathbf{B}}$ is not invertible. But this case is rarely encountered in practical problems, since in practice the number of measurement vectors is generally small, as we explained previously. The presence of noise in practical problems also requires the use of the regularized form (\ref{equ:Brule_lowSNR}), which is always invertible.}. However, in low or medium SNR cases (e.g. $\mathrm{SNR} \leq 20 \mathrm{dB}$) it is not robust due to errors from the estimated $\gamma_i$ and $\mathbf{X}_{i\cdot}$. For these cases, we suggest adding a regularization item in $\widetilde{\mathbf{B}}$, namely,
\begin{eqnarray}
\widetilde{\mathbf{B}} &\leftarrow& \sum_{i=1}^M \frac{\mathbf{X}_{i \cdot}^T \mathbf{X}_{i \cdot} }{\gamma_i} + \eta \mathbf{I}
\label{equ:Brule_lowSNR}
\end{eqnarray}
where $\eta$ is a positive scalar. This regularized form (\ref{equ:Brule_lowSNR}) ensures that $\widetilde{\mathbf{B}}$ is positive definite.

Similarly, we simplify the $\lambda$ learning rule (\ref{equ:lambdaRule}) as follows:
\begin{eqnarray}
\lambda  &\leftarrow& \frac{\|\mathbf{y}-\mathbf{D} \boldsymbol{\mu}_x\|_2^2 + \lambda  \big[ML-\mathrm{Tr}( \mathbf{\Sigma}_x \mathbf{\Sigma}_0^{-1})\big]}{NL} \nonumber \\
&=&  \frac{\|\mathbf{y}-\mathbf{D} \boldsymbol{\mu}_x\|_2^2 + \lambda  \mathrm{Tr}( \mathbf{\Sigma}_0 \mathbf{D}^T \mathbf{\Sigma}_y^{-1}\mathbf{D})}{NL} \label{equ:comment4} \\
& \approx & \frac{1}{NL} \|\mathbf{Y}-\mathbf{\Phi} \mathbf{X} \|_\mathcal{F}^2 + \frac{\lambda}{NL} \mathrm{Tr}\Big[ ( \mathbf{\Gamma} \otimes  \mathbf{B})(\mathbf{\Phi}^T \otimes \mathbf{I}) \nonumber \\
&& \cdot \big( (\lambda \mathbf{I} + \mathbf{\Phi} \mathbf{\Gamma} \mathbf{\Phi}^T )^{-1} \otimes  \mathbf{B}^{-1} \big) (\mathbf{\Phi} \otimes \mathbf{I})   \Big]
\label{equ:comment2} \\
&=& \frac{1}{NL} \|\mathbf{Y}-\mathbf{\Phi} \mathbf{X} \|_\mathcal{F}^2 + \frac{\lambda }{N} \mathrm{Tr}\big[ \mathbf{\Phi} \mathbf{\Gamma} \mathbf{\Phi}^T ( \lambda \mathbf{I} + \mathbf{\Phi} \mathbf{\Gamma} \mathbf{\Phi}^T )^{-1}   \big] \nonumber \\
\label{equ:lambdaRule_simplified}
\end{eqnarray}
where in (\ref{equ:comment4}) we use the first equation in (\ref{equ:Sigma_x}), and in (\ref{equ:comment2}) we use the approximation (\ref{equ:approximationFormula}). Empirically, we find that setting the off-diagonal elements of $\mathbf{\Phi} \mathbf{\Gamma} \mathbf{\Phi}^T$ to zeros further improves the robustness of the $\lambda$ learning rule in strongly noisy cases. In our experiments we will use the modified version when $\mathrm{SNR} \leq 20 \mathrm{dB}$.

We denote the algorithm using the learning rules (\ref{equ:MSBL_cov}), (\ref{equ:X_est_MSBL}), (\ref{equ:gammaRule_tSBL}), (\ref{equ:Brule_simplified}), (\ref{equ:Brule_norm}) (or (\ref{equ:Brule_lowSNR})), and (\ref{equ:lambdaRule_simplified}) by \emph{\textbf{T-MSBL}} (the name emphasizes the algorithm is a \emph{temporal} extension of MSBL). Note that T-MSBL cannot be derived by modifying the cost function of MSBL.

Comparing the $\gamma_i$ learning rule of T-MSBL (Eq.(\ref{equ:gammaRule_tSBL})) with the one of MSBL (Eq.(\ref{equ:gammaRuleInMSBL})),  we observe that the only change is the replacement of $\|  \mathbf{X}_{i \cdot} \|_2^2$ with $ \mathbf{X}_{i \cdot}  \mathbf{B}^{-1}  \mathbf{X}_{i \cdot}^T$, which incorporates the temporal correlation of the sources. Hence, T-MSBL has only extra computational load for calculating the matrix $\mathbf{B}$ and the item $\mathbf{X}_{i\cdot}\mathbf{B}^{-1}\mathbf{X}_{i\cdot}^T$ \footnote{Here we do not compare the $\lambda$ learning rules of both algorithms, since in some cases one can feed the algorithms with suitable fixed values of $\lambda$, instead of using the $\lambda$ learning rules. However, the computational load of the simplified $\lambda$ learning rule of T-MSBL is also not high.}. Since the matrix $\mathbf{B}$ has a small size and is positive definite and symmetric, the extra computational load is low.

Note that $ \mathbf{X}_{i \cdot}  \mathbf{B}^{-1}  \mathbf{X}_{i \cdot}^T$ is the quadratic Mahalanobis distance between $\mathbf{X}_{i \cdot}$ and its mean (a vector of zeros). In the following section we will get more insight into this change.

\section{Analysis of Global Minimum and Local Minima}
\label{sec:analysis}

Since our bSBL framework generalizes the basic SBL framework, many proofs below are rooted in the  theoretic work on the basic SBL \cite{David2004IEEE}. However, some essential modifications are necessary in order to adapt the results to the bSBL model. Due to the equivalence of the original MMV model (\ref{equ:MMV basicmodel}) and the transformed block sparsity model (\ref{equ:blocksparsemodel}), in the following discussions we use  (\ref{equ:MMV basicmodel}) or (\ref{equ:blocksparsemodel}) interchangeably and per convenience.

Throughout our analysis,  the true source matrix is denoted by $\mathbf{X}_{\mathrm{gen}}$, which is the sparsest solution among all the possible solutions. The number of nonzero rows in $\mathbf{X}_{\mathrm{gen}}$ is denoted by $K_0$. We assume that $\mathbf{X}_{\mathrm{gen}}$ is full column-rank, the dictionary matrix $\mathbf{\Phi}$ satisfies the URP condition \cite{Gorodnitsky1997}, and the matrix $\mathbf{B}$ (or $\mathbf{B}_i,\forall i$) and its estimate are positive definite.

\subsection{Analysis of the Global Minimum}

We have the following result on the global minimum of the cost function (\ref{equ:costfunc}) \footnote{For convenience, in this theorem we consider the cost function with $\mathbf{\Sigma}_0$ given by (\ref{equ:Sigma0}), i.e. the one before we use our strategy to avoid the overfitting.}:

\theorem{In the limit as $\lambda \rightarrow 0$, assuming $K_0 < (N+L)/2$ , for the cost function (\ref{equ:costfunc}) the unique global minimum $\widehat{\boldsymbol{\gamma}} \triangleq [\widehat{\gamma}_1,\cdots,\widehat{\gamma}_M]$ produces a source estimate $\widehat{\mathbf{X}}$  that equals to $\mathbf{X}_{\mathrm{gen}}$ irrespective of the estimated $\widehat{\mathbf{B}}_i, \, \forall i$, where $\widehat{\mathbf{X}}$ is obtained from $\mathrm{vec}(\widehat{\mathbf{X}}^T) = \widehat{\mathbf{x}} $ and $\widehat{\mathbf{x}}$ is computed using Eq.(\ref{equ:noisySolution}).\label{theorem:global}}

The proof is given in the Appendix.

If we change the condition $K_0 < (N+L)/2$ to $K_0 < N$, then we  have the conclusion that the source estimate $\widehat{\mathbf{X}}$ equals to $\mathbf{X}_{\mathrm{gen}}$ with probability 1, irrespective of  $\widehat{\mathbf{B}}_i, \, \forall i$. This is due to the result in \cite{Elad2006} that if $K_0 < N$ the above conclusion still holds for all $\mathbf{X}$ except on a set with zero measure.

Note that  $\widehat{\boldsymbol{\gamma}}$ is a function of the estimated $\widehat{\mathbf{B}}_i$ ($\forall i$). However, the theorem implies that even when the estimated $\widehat{\mathbf{B}}_i$ is different from the true $\mathbf{B}_i$, the estimated sources are the true sources at the global minimum of the cost function.  As a reminder,  in deriving our algorithms, we assumed $\mathbf{B}_i = \mathbf{B}$ ($\forall i$) to avoid overfitting. Theorem \ref{theorem:global} ensures our algorithms using this strategy also have the global minimum property. Also, the theorem explains why  MSBL has the ability to exactly recover true sources in noiseless cases even when  sources  are temporally correlated. But we hasten to add that this does not mean $\mathbf{B}$ is not important for the performance  of the algorithms. For instance, MSBL is more frequently attracted to local minima than our proposed algorithms, as experiments show later.

\subsection{Analysis of the Local Minima}

In this subsection we discuss the local minimum property of the cost function $\mathcal{L}$ in (\ref{equ:costfunc}) with respect to $\boldsymbol{\gamma} \triangleq [\gamma_1,\cdots,\gamma_M]$, in which $\mathbf{\Sigma}_0 = \mathbf{\Gamma} \otimes \mathbf{B}$ for fixed  $\mathbf{B}$. Before presenting our results, we provide two lemmas needed to prove the results.

\lemma{$\log|\mathbf{\Sigma}_y| \triangleq \log |\lambda \mathbf{I} + \mathbf{D} \mathbf{\Sigma}_0 \mathbf{D}^T|$ is concave with respect to  $\boldsymbol{\gamma}$.\label{lemma:concavity}}

This can be shown using the composition property of concave functions \cite{BoydBook}.

\lemma{$\mathbf{y}^T \mathbf{\Sigma}_y^{-1} \mathbf{y}$ equals a constant $C$ when $\mathbf{\boldsymbol{\gamma}}$  satisfies the linear constraints
\begin{eqnarray}
\mathbf{A} \cdot (\mathbf{\boldsymbol{\gamma}} \otimes \mathbf{1}_L) = \mathbf{b}
\label{equ:b_A_gamma}
\end{eqnarray}
with
\begin{eqnarray}
\mathbf{b} &\triangleq& \mathbf{y} - \lambda \mathbf{u}   \label{equ:Lemma2_b} \\
\mathbf{A} &\triangleq& (\mathbf{\Phi} \otimes \mathbf{B}) \mathrm{diag}(\mathbf{D}^T \mathbf{u})  \label{equ:Lemma2_A}
\end{eqnarray}
where $\mathbf{A}$ is full row rank,  $\mathbf{1}_L$ is an $L \times 1$ vector of ones, and $\mathbf{u}$ is any fixed vector such that $\mathbf{y}^T \mathbf{u} = C$.\label{lemma:b_AGamma}}

The proof is given in the Appendix. According to the definition of basic feasible solution (BFS) \cite{LuenbergerBook}, we know that if $\boldsymbol{\gamma}$ satisfies Equation (\ref{equ:b_A_gamma}), then it is a BFS to (\ref{equ:b_A_gamma}) if $\| \boldsymbol{\gamma} \|_0 \leq NL$, or a degenerate BFS to (\ref{equ:b_A_gamma}) if $\| \boldsymbol{\gamma} \|_0 < NL$. Now we give the following result:

\theorem{Every local minimum of the cost function $\mathcal{L}$ with respect to $\boldsymbol{\gamma}$ is achieved at a solution with $\|\widehat{\boldsymbol{\gamma}} \|_0 \leq NL$, regardless of the values of $\lambda$ and $\mathbf{B}$. \label{theorem:local}}

The proof is given in the Appendix.

Admittedly, the bound on the local minima $\|\widehat{\boldsymbol{\gamma}} \|_0$ is loose, and it is not meaningful when $NL > M$. However, we empirically found that $\|\widehat{\boldsymbol{\gamma}} \|_0$ actually is very smaller than $NL$.

Now, we calculate the local minima of the cost function $\mathcal{L}$. The result can provide some insights to the role of $\mathbf{B}$. Particularly, we are more interested in the local minima satisfying $\|\widehat{\boldsymbol{\gamma}} \|_0 \leq N$, since the global minimum satisfies $\|\widehat{\boldsymbol{\gamma}} \|_0 < N$. For these local minima, we have the following result:

\lemma{In noiseless cases ($\lambda \rightarrow 0$), for every local minimum of $\mathcal{L}$ that satisfies $\|\widehat{\boldsymbol{\gamma}} \|_0 \triangleq K\leq N$, its $i$-th nonzero element is given by $\widehat{\gamma}_{(i)} = \frac{1}{L} \widetilde{\mathbf{X}}_{i\cdot} {\mathbf{B}}^{-1} \widetilde{\mathbf{X}}^T_{i\cdot} \,(i=1,\cdots,K)$, where $\widetilde{\mathbf{X}}_{i\cdot}$ is the $i$-th nonzero row of $\widehat{\mathbf{X}}$ and  $\widehat{\mathbf{X}}$ is the basic feasible solution to $\mathbf{Y}=\mathbf{\Phi}\mathbf{X}$. \label{lemma:localSolution}}

The proof is given in the Appendix.

From this lemma we immediately have the closed form of the global minimum.

$\mathbf{B}$ actually plays a role of temporally whitening the sources during the learning of $\boldsymbol{\gamma}$. To see this, assume all the sources have the same correlation structure, i.e. share the same matrix $\mathbf{B}$. Let $\mathbf{Z}_{i\cdot} \triangleq \widetilde{\mathbf{X}}_{i\cdot} \mathbf{B}^{-1/2}$. From Lemma \ref{lemma:localSolution}, at the global minimum we have $\widehat{\gamma}_{(i)} = \frac{1}{L} \mathbf{Z}_{i\cdot}  \mathbf{Z}^T_{i\cdot} \,(i=1,\cdots,K_0)$. On the other hand, in the case of i.i.d. sources, at the global minimum we have $\widehat{\gamma}_{(i)} = \frac{1}{L} \widetilde{\mathbf{X}}_{i\cdot}  \widetilde{\mathbf{X}}^T_{i\cdot} \,(i=1,\cdots,K_0)$. So the results for the two cases have the same form. Since $E\{\mathbf{Z}_{i\cdot}^T \mathbf{Z}_{i\cdot} \} = \gamma_i \mathbf{I}$, we can see in the learning of $\boldsymbol{\gamma}$, $\mathbf{B}$ plays the role of whitening each source. This gives us a motivation to modify most state-of-the-art iterative reweighted algorithms by temporally whitening the estimated sources during iterations \cite{Zhilin_ICASSP2011,Zhilin_ICML2011}.

\section{Computer Experiments}
\label{sec:experiment}

Extensive computer experiments have been conducted and a few representative and informative results are presented.
All the experiments consisted of 1000 independent trials. In each trial a dictionary matrix  $\mathbf{\Phi} \in \mathbb{R}^{N \times M}$ was created with columns uniformly drawn from the surface of a unit hypersphere (except the experiment in Section \ref{exp:extremeEg}), as advocated by Donoho et al \cite{Donoho2004}. And the source matrix $\mathbf{X}_{\mathrm{gen}} \in \mathbb{R}^{M \times L}$ was randomly generated with $K$ nonzero rows (i.e. sources). In each trial the indexes of the sources were randomly chosen. In most experiments (except to the  experiment in Section \ref{exp:varyARp}) each source was generated as AR(1) process. Thus the AR coefficient of the $i$-th source, denoted by $\beta_i$, indicated its temporal correlation. As done in \cite{Tropp2006SP1,Bach2008}, for noiseless cases, the $\ell_2$ norm of each source was rescaled to be uniformly distributed between $1/3$ and 1; for noisy cases, rescaled to be unit norm. Finally, the measurement matrix $\mathbf{Y}$ was constructed by $\mathbf{Y} = \mathbf{\Phi} \mathbf{X}_{\mathrm{gen}} + \mathbf{V}$ where $\mathbf{V}$ was a zero-mean homoscedastic Gaussian noise matrix with variance adjusted to have a desired value of SNR, which is defined by $\mathrm{SNR}(\mathrm{dB}) \triangleq 20\log_{10} (\|\mathbf{\Phi} \mathbf{X}_{\mathrm{gen}}\|_\mathcal{F} / \| \mathbf{V} \|_\mathcal{F} )$.

We used two performance measures. One was the \emph{Failure Rate} defined in \cite{David2007IEEE}, which indicated the percentage of failed trials in the total trials. In  noiseless cases, a failed trial was recognized if the indexes of estimated sources  were not the same as the true indexes. In noisy cases, since any algorithm cannot recover $\mathbf{X}_{\mathrm{gen}}$ exactly in these cases, a failed trial was recognized if the indexes of estimated sources with the $K$ largest $\ell_2$ norms were not the same as the true indexes. In noisy cases, the \emph{mean square error} (MSE) was also used as a performance measure, defined by $\| \widehat{\mathbf{X}} - \mathbf{X}_{\mathrm{gen}}  \|_\mathcal{F}^2/ \| \mathbf{X}_{\mathrm{gen}} \|_\mathcal{F}^2$, where $\widehat{\mathbf{X}}$ was the estimated source matrix.

In our experiments we compared our T-SBL and T-MSBL with the following algorithms:
\begin{itemize}
  \item MSBL, proposed in \cite{David2007IEEE} \footnote{The MATLAB code was downloaded at \url{http://dsp.ucsd.edu/~zhilin/MSBL_code.zip}.};

  \item MFOCUSS, the regularized M-FOCUSS proposed in \cite{Cotter2005}. In all the experiments, we set its p-norm $p=0.8$, as suggested by the authors \footnote{The MATLAB code was downloaded at \url{http://dsp.ucsd.edu/~zhilin/MFOCUSS.m}.};

  \item SOB-MFOCUSS, a smoothness constrained M-FOCUSS proposed in \cite{Cichocki2008}. In all the experiments, we set its p-norm $p=0.8$. For its smoothness matrix,  we chose the identity matrix when $L\leq 2$, and a second-order smoothness matrix when $L \geq 3$, as suggested by the authors. Since in our experiments $L$ is small, no overlap blocks were used \footnote{The MATLAB code was provided by the first author of \cite{Cichocki2008} in personal communication. In the code the second-order smoothness matrix $\mathbf{S}$ was defined as (in MATLAB notations): $\mathbf{S} = \mathrm{eye}(L) - 0.25*(\mathrm{diag}(\mathbf{e}(1:L-1),-1) + \mathrm{diag}(\mathbf{e}(1:L-1),1) + (\mathrm{diag}(\mathbf{e}(1:L-2),-2) + \mathrm{diag}(\mathbf{e}(1:L-2),2)))$, where $\mathbf{e}$ is an $L \times 1$ vector with ones.};

  \item ISL0, an improved smooth $\ell_0$ algorithm for the MMV model which was proposed in \cite{Hyder2010b}. The regularization parameters were chosen according to the authors' suggestions \footnote{The MATLAB code was provided by the first author of \cite{Hyder2010b} in personal communication.};

  \item Reweighted $\ell_1/\ell_2$, an iterative reweighted $\ell_1/\ell_2$ algorithm  suggested in \cite{David2010reweighting}. It is an MMV extension of the iterative reweighted $\ell_1$ algorithm \cite{Candes2008reweighting} via the mixed $\ell_1/\ell_2$ norm. The algorithm is given by
            \begin{enumerate}
                \item Set the iteration count $k$ to zero and $w_i^{(0)}=1,i=1,\cdots,M$
                \item Solve the weighted MMV $\ell_1$ minimization problem
                    \begin{eqnarray}
                    \mathbf{X}^{(k)} = \arg\min \sum_{i=1}^M  w_i^{(k)} \| \mathbf{X}_{i\cdot} \|_2  \quad \mathrm{s.t.}\; \mathbf{Y}=\mathbf{\Phi} \mathbf{X}
                    \nonumber
                    \end{eqnarray}
                \item Update the weights for each $i=1,\cdots,M$
                    \begin{eqnarray}
                    w_i^{(k+1)} = \frac{1}{\| \mathbf{X}_{i\cdot}^{(k)} \|_2 + \epsilon^{(k)}}  \nonumber
                    \end{eqnarray}
                    where $\epsilon^{(k)}$ is adaptively adjusted as in \cite{Candes2008reweighting};
                \item Terminate on convergence or when $k$ attains a specified maximum number of iterations $k_{\mathrm{max}}$. Otherwise, increment $k$ and go to Step 2).
            \end{enumerate}
       For  noisy cases, Step 2) is modified to
            \begin{eqnarray}
            \mathbf{X}^{(k)} = \arg\min \sum_{i=1}^M  w_i^{(k)} \| \mathbf{X}_{i\cdot} \|_2  \quad \mathrm{s.t.}\; \|\mathbf{Y}-\mathbf{\Phi} \mathbf{X}\|_\mathcal{F} \leq \delta  \nonumber
            \end{eqnarray}
        Throughout our experiments, $k_{\mathrm{max}}=5$. We implemented it using the CVX optimization toolbox \footnote{The toolbox was downloaded at: \url{http://cvxr.com/cvx/}}.
\end{itemize}
In noisy cases, we chose the optimal values for  the regularization parameter $\lambda$ in MFOCUSS and the parameter $\delta$ in Reweighted $\ell_1/\ell_2$ by exhaustive search. Practically, we used a set of candidate parameter values and for each value we ran an algorithm for 50 trials, and then picked up the one which gave the smallest averaged failure rate. By comparing enough number of candidate values we could ensure a nearly optimal value of the regularization parameter for this algorithm. For T-MSBL, T-SBL and MSBL, we fixed  $\lambda=10^{-9}$ for noiseless cases, and used their $\lambda$ learning rules for noisy cases. Besides, for T-MSBL we chose the learning rule (\ref{equ:Brule_lowSNR}) with $\eta = 2$ to estimate $\mathbf{B}$ when $\mathrm{SNR} \leq 15 \mathrm{dB}$.

For reproducibility, the experiment codes can be downloaded at \url{http://dsp.ucsd.edu/~zhilin/TSBL_code.zip}.

\subsection{Benefit from Multiple Measurement Vectors at Different Temporal Correlation Levels}
\label{exp:varyL}

In this experiment we study how algorithms benefit from multiple measurement vectors and how the benefit is discounted by the temporal correlation of sources. The dictionary matrix $\mathbf{\Phi}$ was of the size $25 \times 125$ and the number of sources $K=12$. The number of measurement vectors $L$ varied from 1 to 4. No noise was added. All the sources were AR(1) processes with the common AR coefficient $\beta$, such that we could easily observe the relationship between  temporal correlation and algorithm performance. Note that for small $L$, modeling sources as AR(1) processes, instead of $\mathrm{AR}(p)$ processes with $p > 1$, is sufficient to cover wide ranges of temporal structure. We compared algorithms at six different temporal correlation levels, i.e. $\beta = -0.9,\,-0.5,\,0,\,0.5,\,0.9,\,0.99$.

\begin{figure}[htb]
\begin{minipage}[b]{.48\linewidth}
  \centering
  \centerline{\epsfig{figure=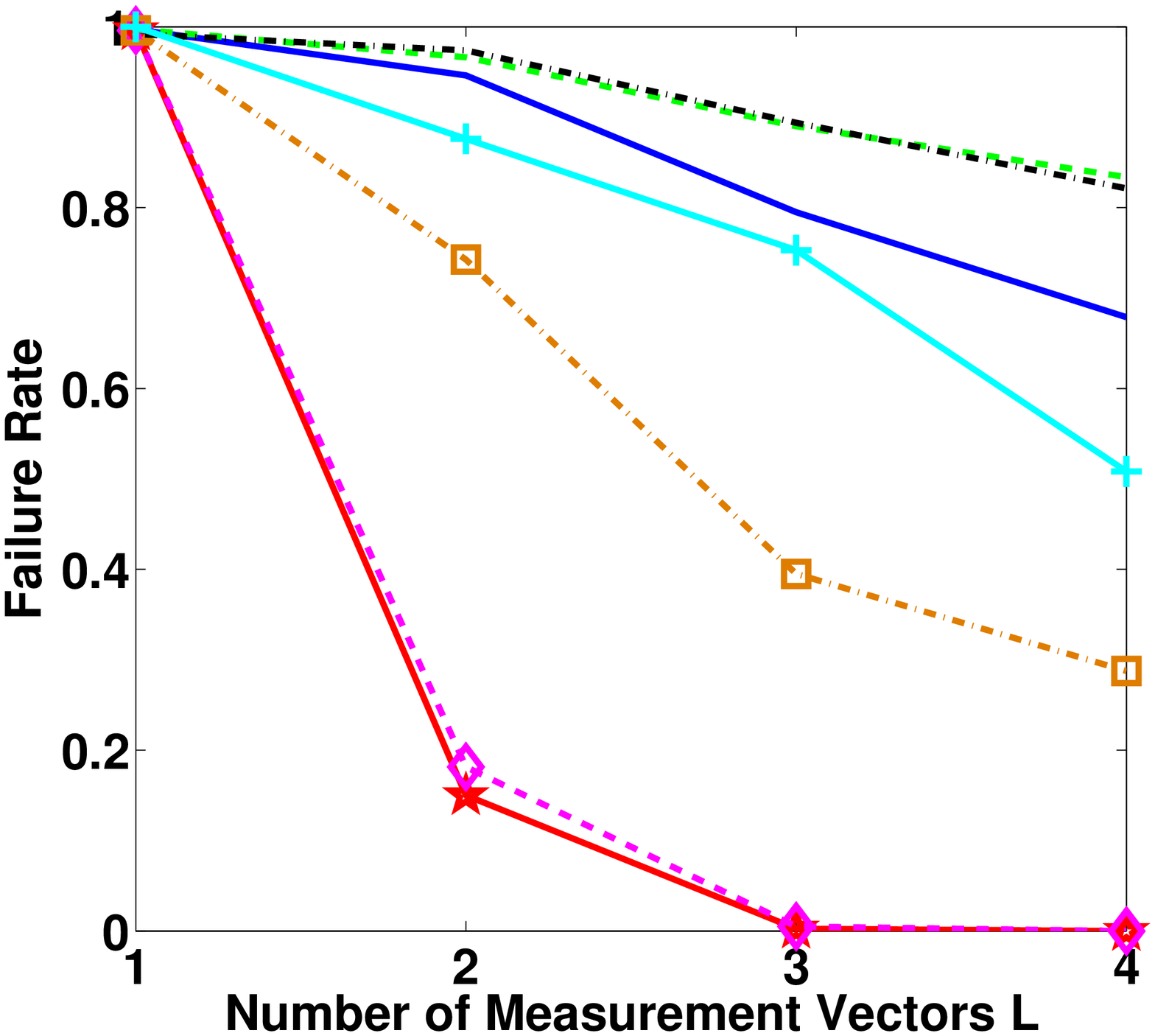,width=4.5cm}}
  \centerline{\footnotesize{(a) $\beta=-0.9$}}
\end{minipage}
\hfill
\begin{minipage}[b]{.48\linewidth}
  \centering
  \centerline{\epsfig{figure=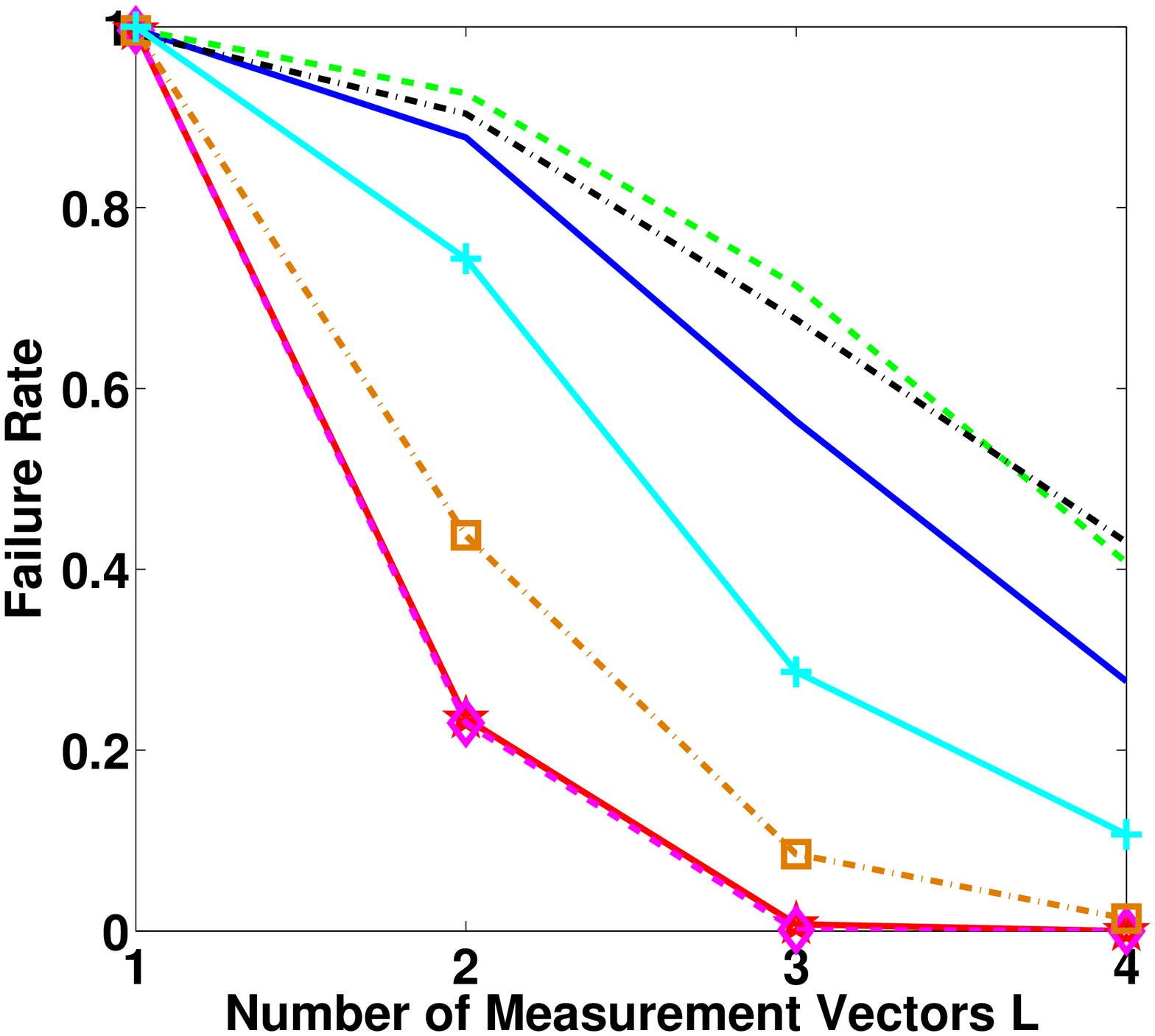,width=4.5cm}}
  \centerline{\footnotesize{(b) $\beta=-0.5$}}
\end{minipage}
\hfill
\begin{minipage}[b]{.48\linewidth}
  \centering
  \centerline{\epsfig{figure=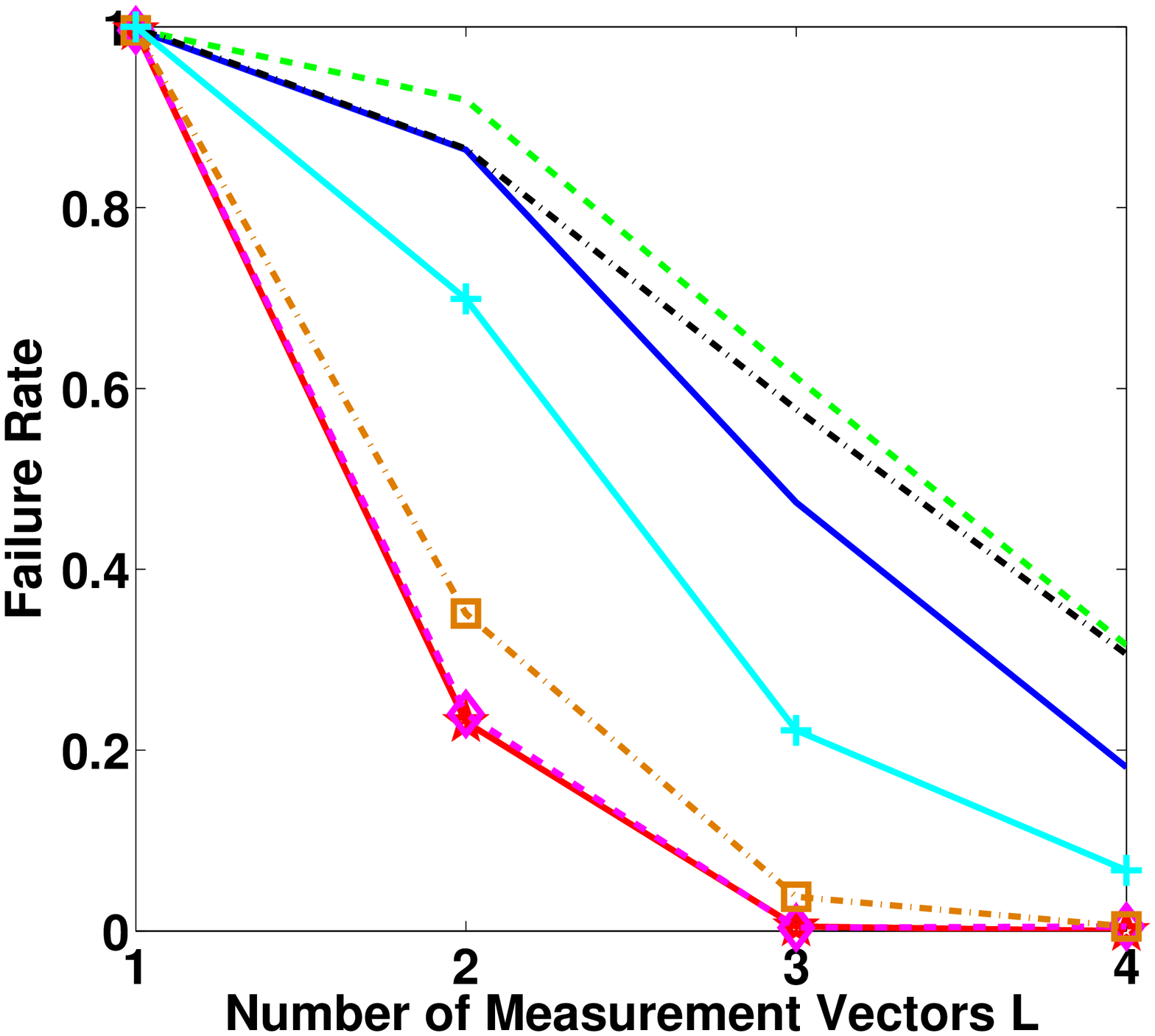,width=4.5cm}}
  \centerline{\footnotesize{(c) $\beta=0$}}
\end{minipage}
\hfill
\begin{minipage}[b]{0.48\linewidth}
  \centering
  \centerline{\epsfig{figure=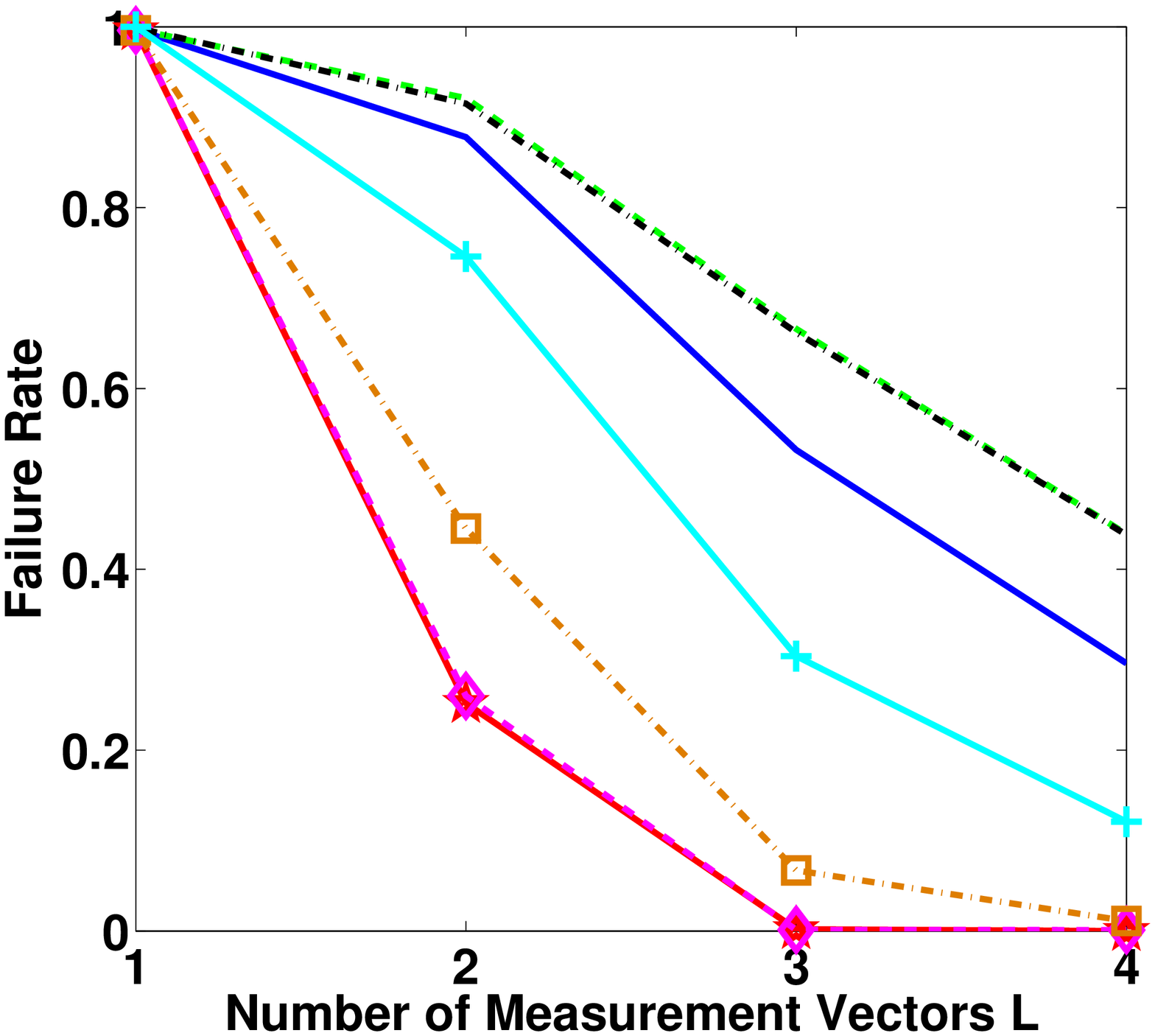,width=4.5cm}}
  \centerline{\footnotesize{(d) $\beta=0.5$}}
\end{minipage}
\hfill
\begin{minipage}[b]{0.48\linewidth}
  \centering
  \centerline{\epsfig{figure=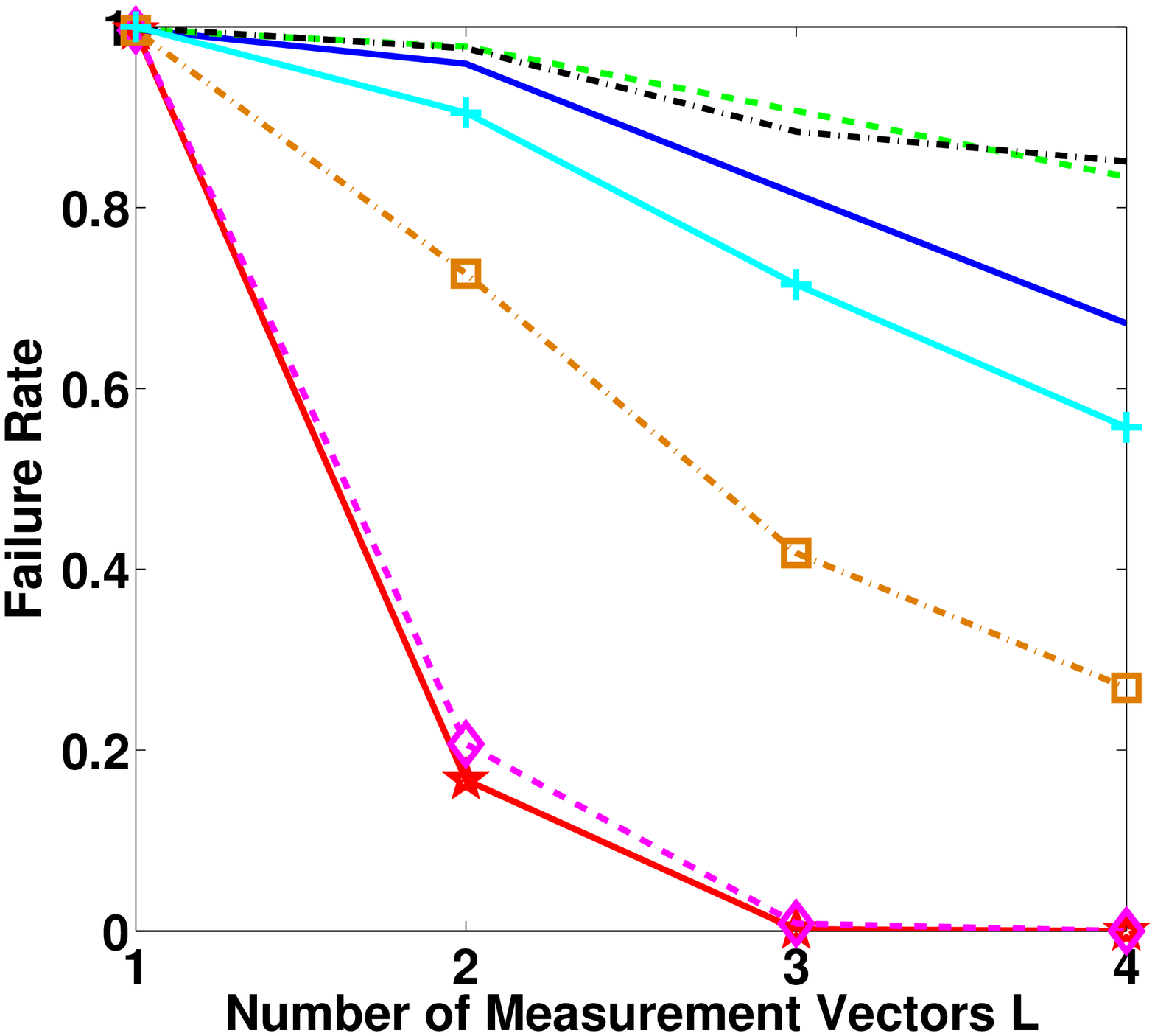,width=4.5cm}}
  \centerline{\footnotesize{(e) $\beta=0.9$}}
\end{minipage}
\hfill
\begin{minipage}[b]{0.48\linewidth}
  \centering
  \centerline{\epsfig{figure=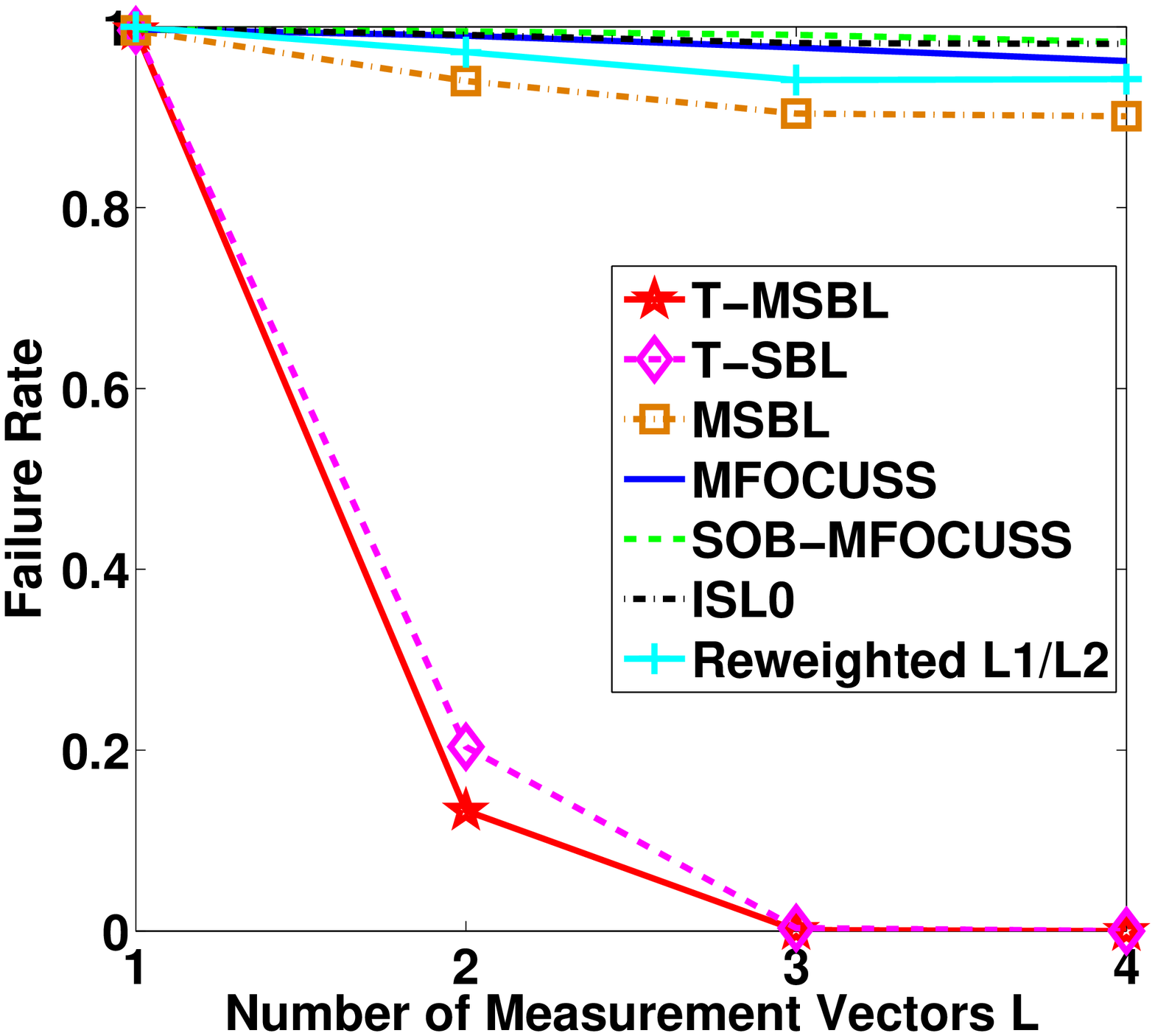,width=4.5cm}}
  \centerline{\footnotesize{(f) $\beta=0.99$}}
\end{minipage}
\caption{Performance of all the algorithms at different temporal correlation levels when $L$ varied from 1 to 4.  }
\label{fig:Simulation_varyL}
\end{figure}

Figure \ref{fig:Simulation_varyL} shows that with $L$ increasing, all the algorithms had better performance. But as $|\beta| \rightarrow 1$, for all the compared algorithms the benefit from multiple measurement vectors diminished. One surprising observation is that our T-MSBL and T-SBL had excellent performance in all cases, no matter what the temporal correlation was. Notice that even sources had no temporal correlation ($\beta=0$), T-MSBL and T-SBL still had better performance than MSBL.

\begin{figure}[htb]
\begin{minipage}[b]{.48\linewidth}
  \centering
  \centerline{\epsfig{figure=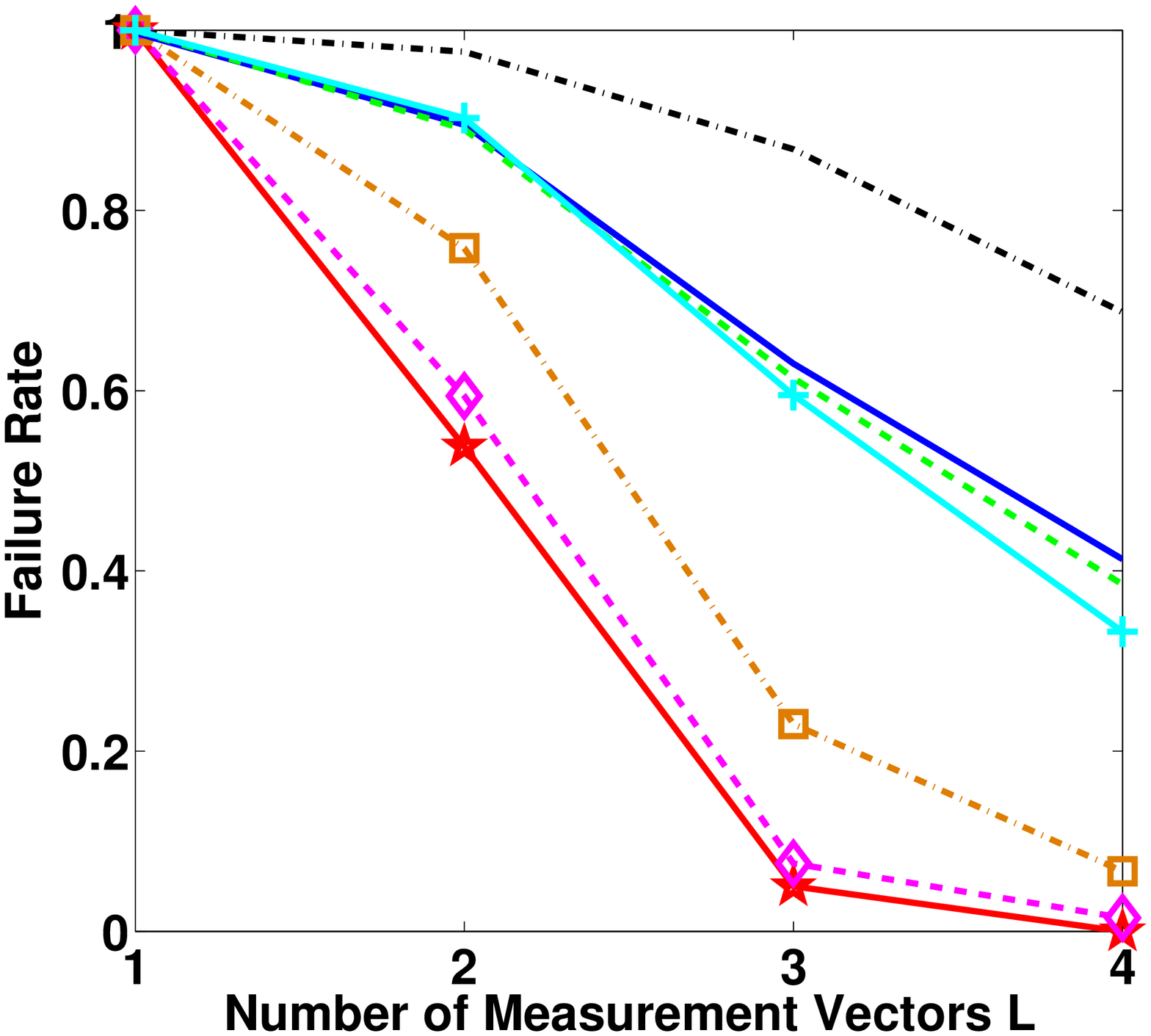,width=4.5cm}}
  \centerline{\footnotesize{(a) $\beta = 0.7$}}
\end{minipage}
\hfill
\begin{minipage}[b]{0.48\linewidth}
  \centering
  \centerline{\epsfig{figure=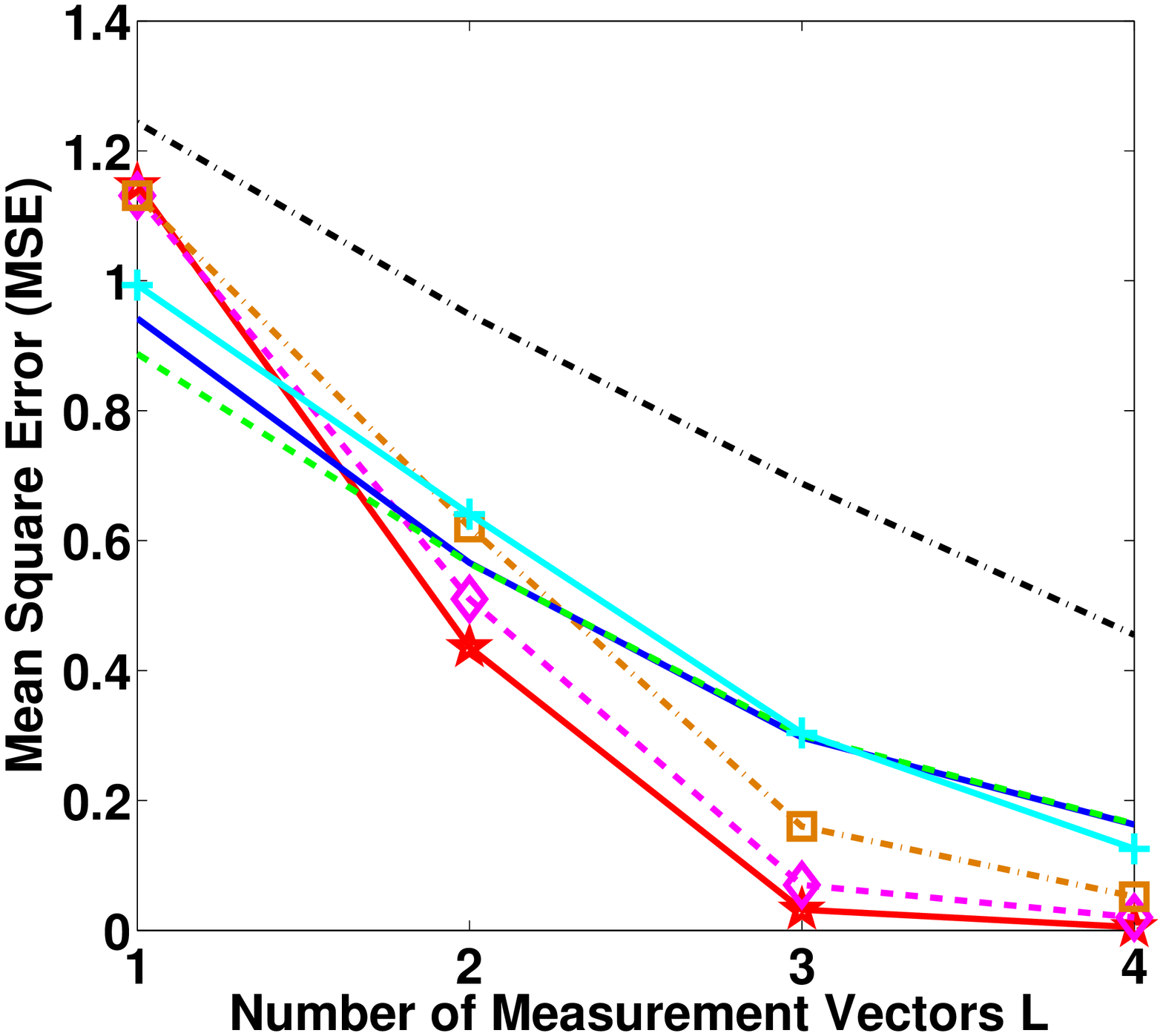,width=4.5cm}}
  \centerline{\footnotesize{(b) $\beta = 0.7$}}
\end{minipage}
\hfill
\begin{minipage}[b]{.48\linewidth}
  \centering
  \centerline{\epsfig{figure=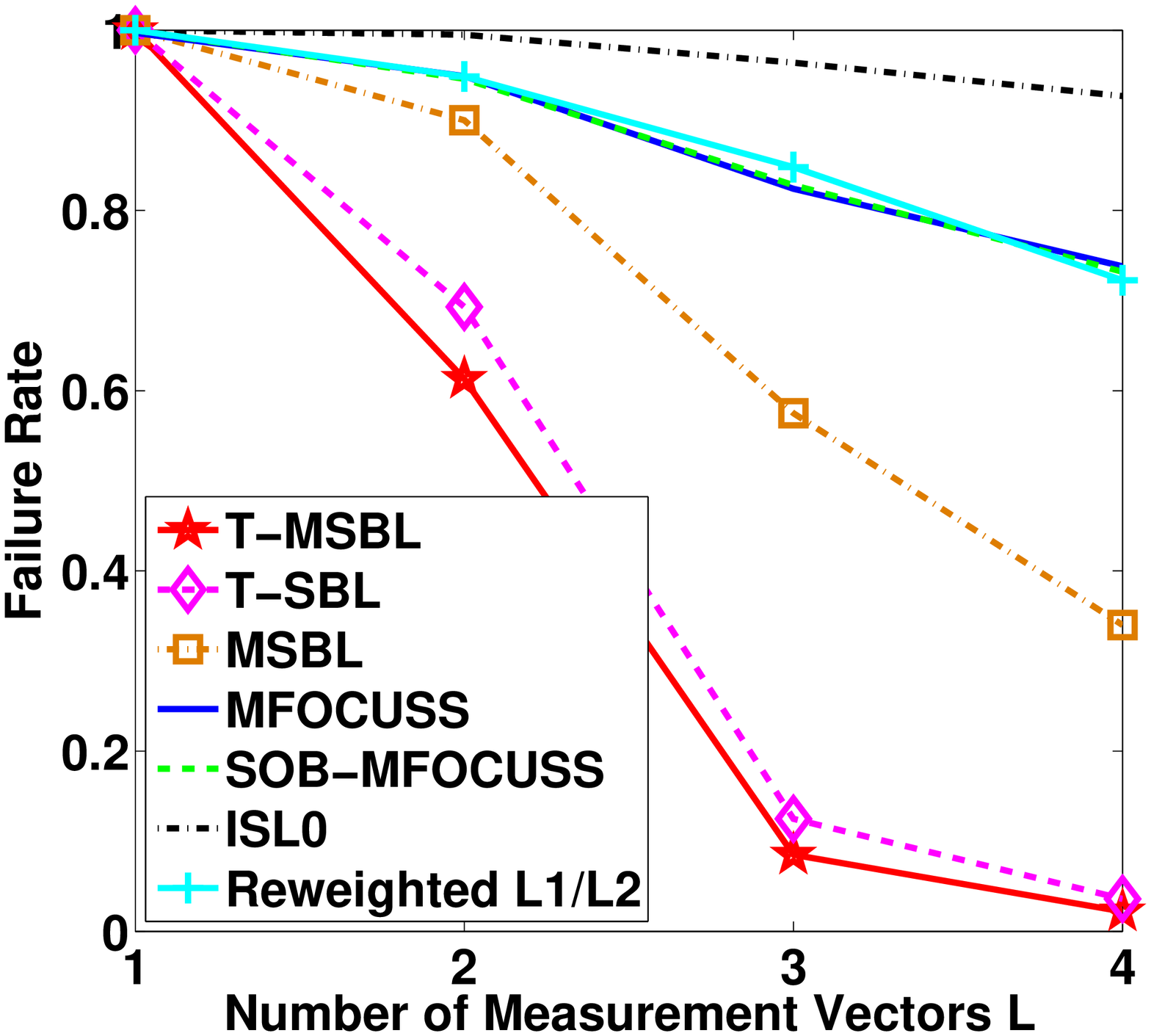,width=4.5cm}}
  \centerline{\footnotesize{(c) $\beta = 0.9$}}
\end{minipage}
\hfill
\begin{minipage}[b]{0.48\linewidth}
  \centering
  \centerline{\epsfig{figure=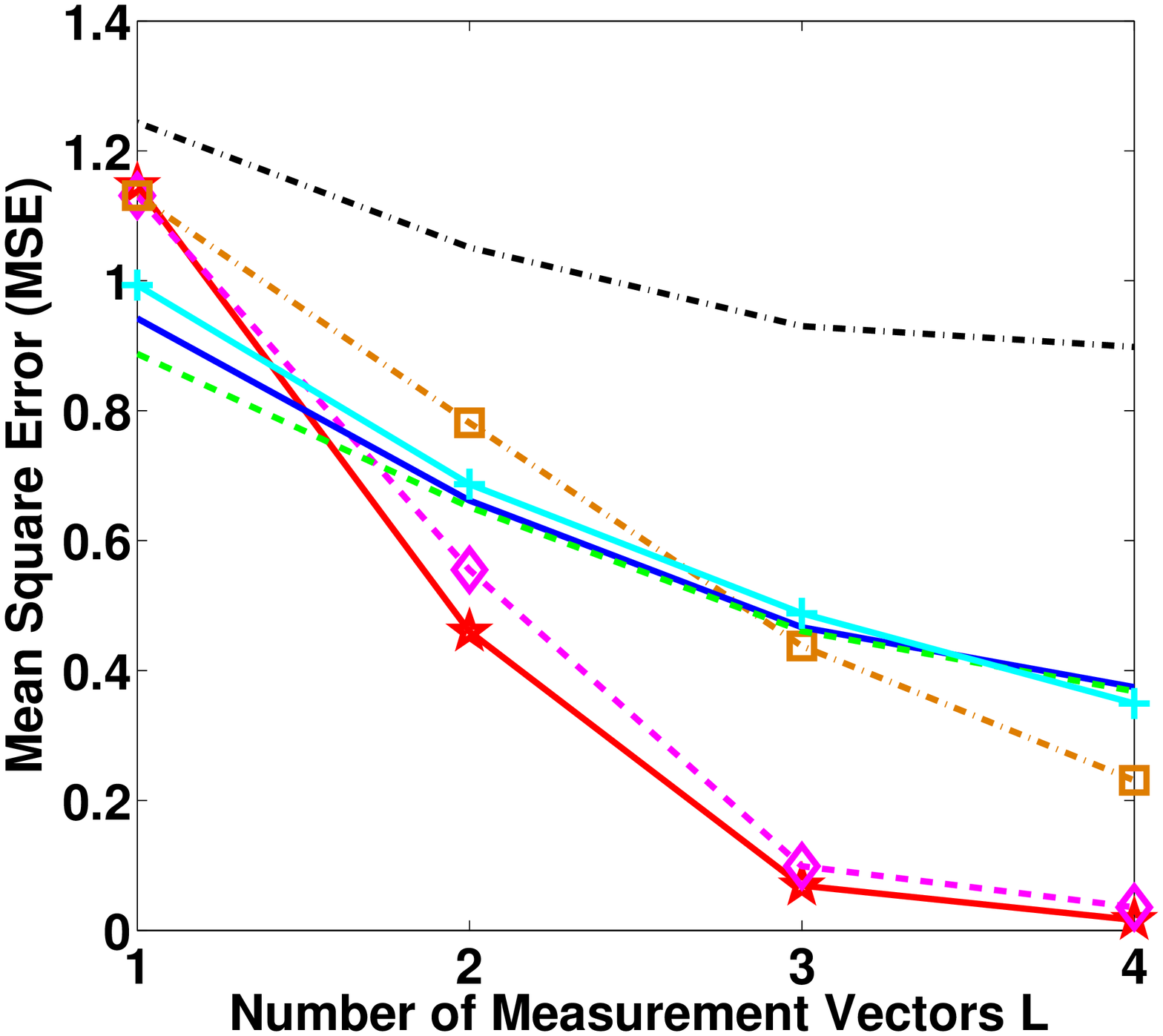,width=4.5cm}}
  \centerline{\footnotesize{(d) $\beta = 0.9$}}
\end{minipage}
\caption{Performance of all the algorithms at different temporal correlation levels when $L$ varied from 1 to 4 and SNR was 25 dB.}
\label{fig:Simulation_varyLnoisy}
\end{figure}

Next we compare all the algorithms in noisy environments. We set $\mathrm{SNR}=25\mathrm{dB}$ while  kept other experimental settings unchanged. The behaviors of all the algorithms were similar to the noiseless case. To save space, we only present the cases of $\beta=0.7$ and $\beta=0.9$ in Fig.\ref{fig:Simulation_varyLnoisy}.

Since the performance of all the algorithms at a given correlation level $\beta$ is the same as their performance at the correlation level $-\beta$, in the following we mainly show their performance at positive correlation levels.

\subsection{Recovered Source Number at Different Temporal Correlation Levels}
\label{exp:varyD}

\begin{figure}[h]
\begin{minipage}[b]{.48\linewidth}
  \centering
  \centerline{\epsfig{figure=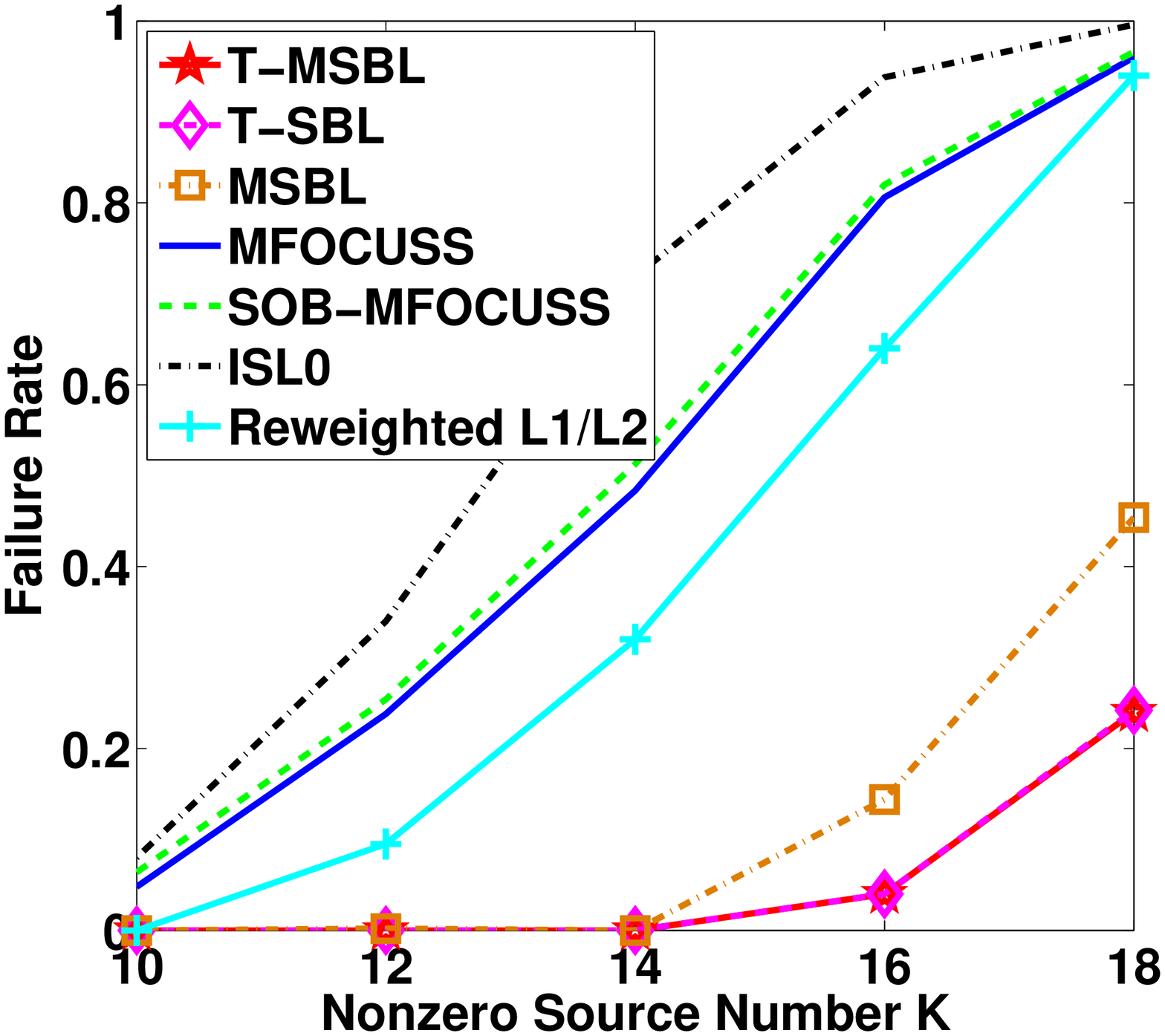,width=4.5cm}}
  \centerline{\footnotesize{(a) $\beta=0$}}
\end{minipage}
\hfill
\begin{minipage}[b]{0.48\linewidth}
  \centering
  \centerline{\epsfig{figure=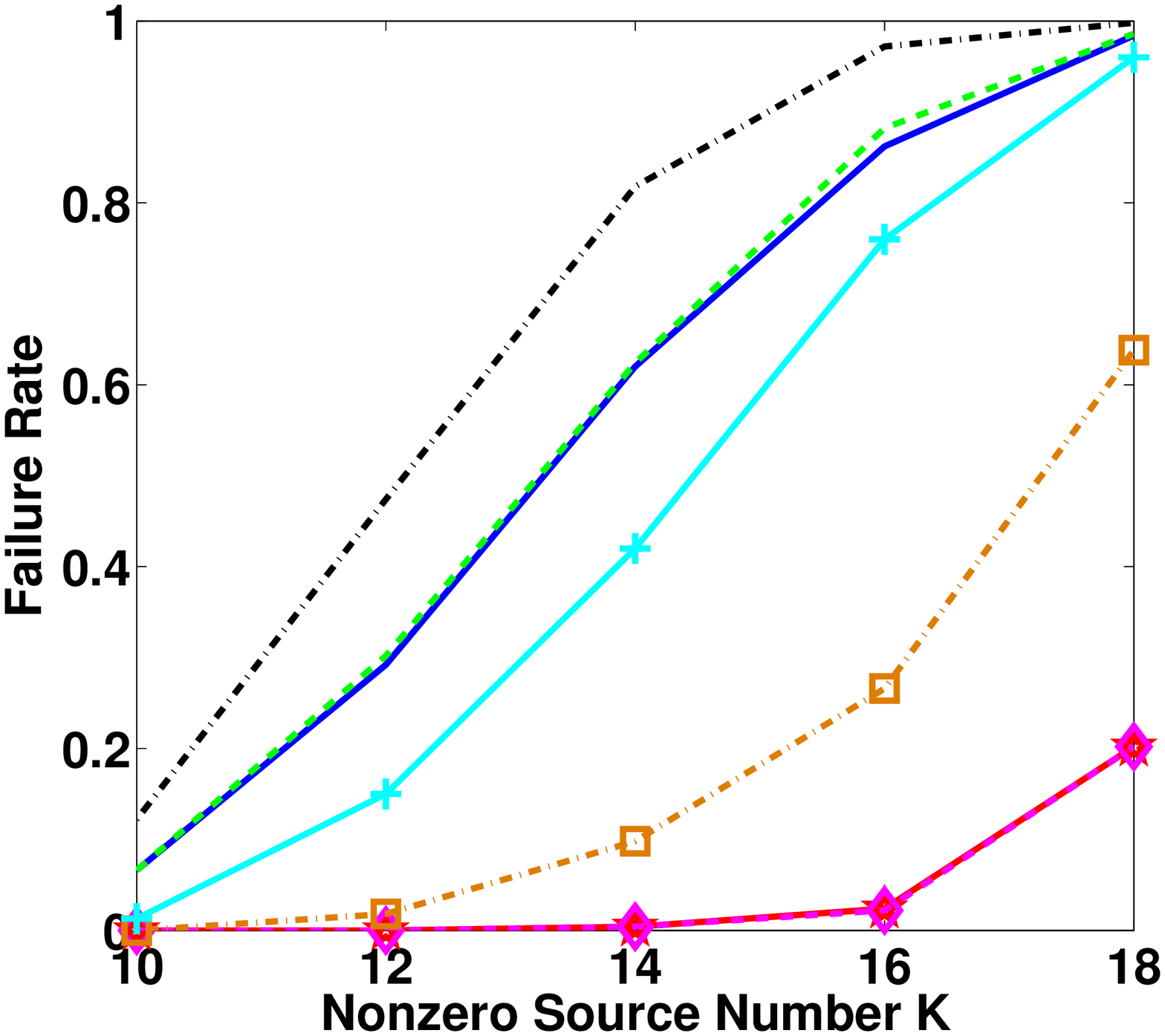,width=4.5cm}}
  \centerline{\footnotesize{(b) $\beta=0.5$}}
\end{minipage}
\hfill
\begin{minipage}[b]{0.48\linewidth}
  \centering
  \centerline{\epsfig{figure=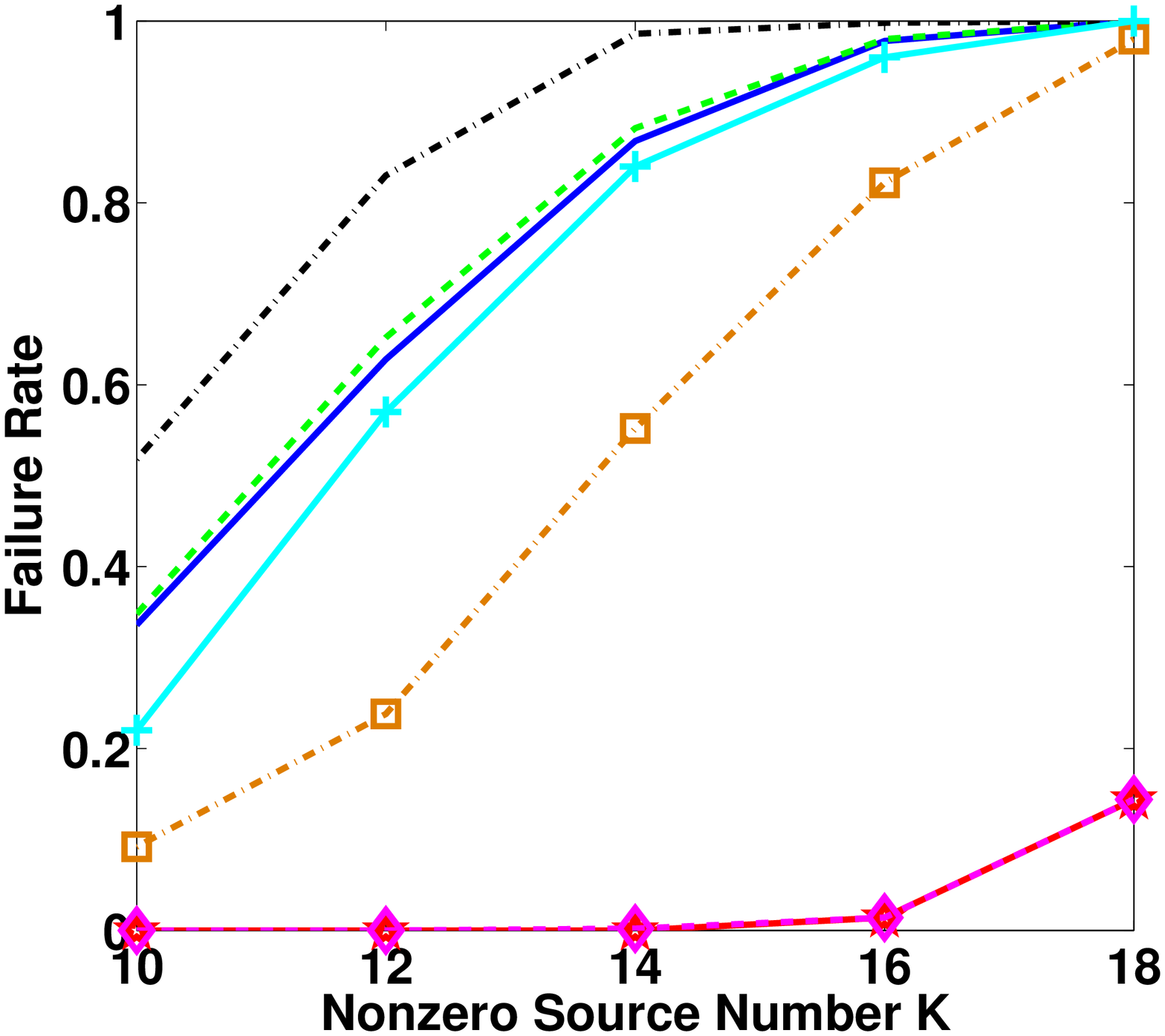,width=4.5cm}}
  \centerline{\footnotesize{(c) $\beta=0.9$}}
\end{minipage}
\hfill
\begin{minipage}[b]{0.48\linewidth}
  \centering
  \centerline{\epsfig{figure=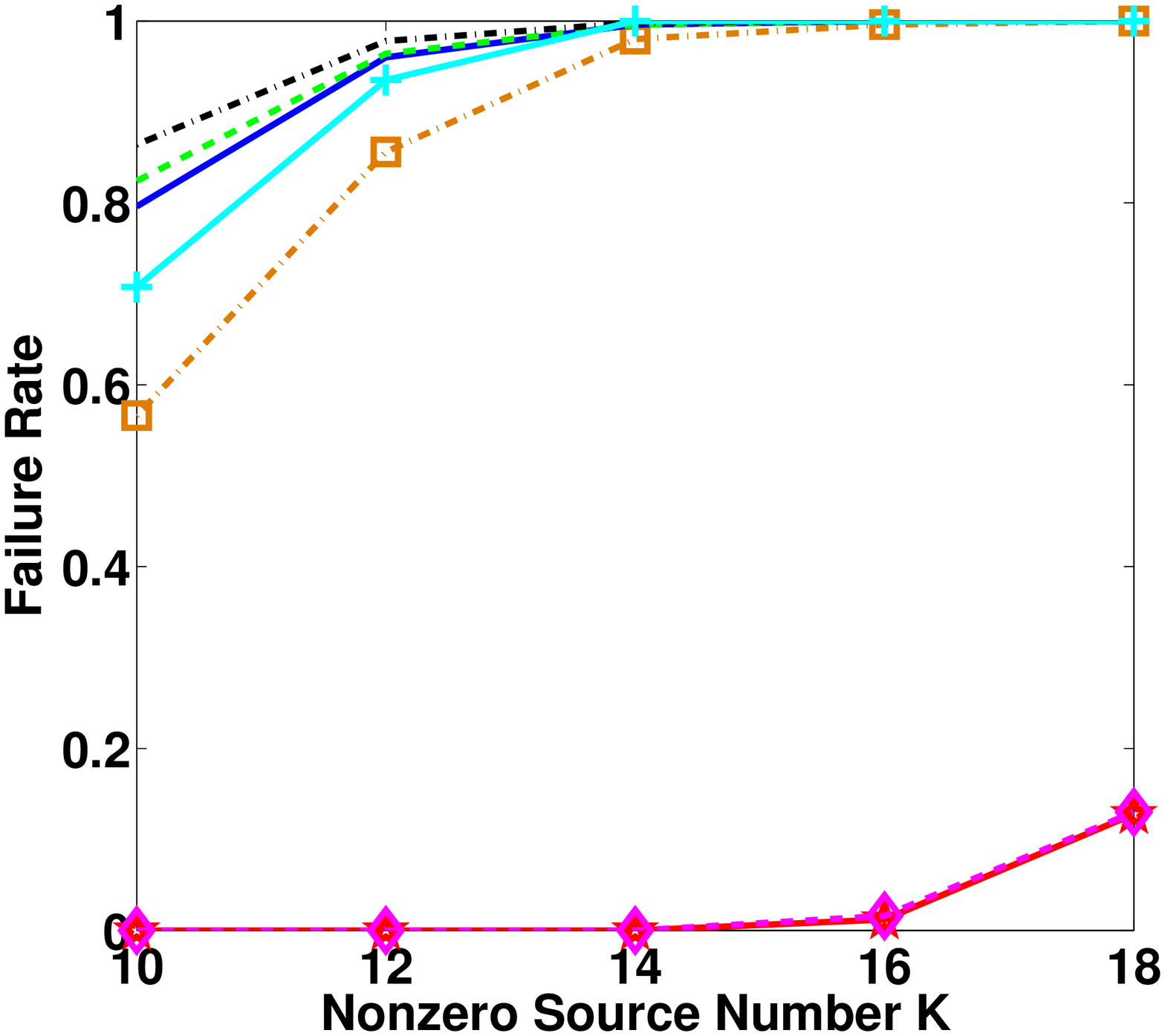,width=4.5cm}}
  \centerline{\footnotesize{(d) $\beta=0.99$}}
\end{minipage}
\caption{Failure rates of all the  algorithms when $K$ varied from 10 to 18 at different temporal correlation levels.}
\label{fig:Simulation_varyD}
\end{figure}

In this experiment we study the effects of temporal correlation on the number of accurately recovered sources in a noiseless case. The dictionary matrix $\mathbf{\Phi}$ was of the size $25 \times 125$. $L$ was 4. $K$ varied from 10 to 18. The sources were generated in the same manner as before. Algorithms were compared at four different temporal correlation levels, i.e. $\beta = 0$, $0.5$, $0.9$, and $0.99$. Results (Fig.\ref{fig:Simulation_varyD}) show that T-MSBL and T-SBL  accurately recovered much more sources than other algorithms, especially at high temporal correlation levels. This indicates that our proposed algorithms are very advantageous in the cases when the source number is large.

\subsection{Ability to Handle Highly Underdetermined Problem}
\label{exp:varyR}

\begin{figure}
\centering
\includegraphics[width=8cm,height=6.5cm]{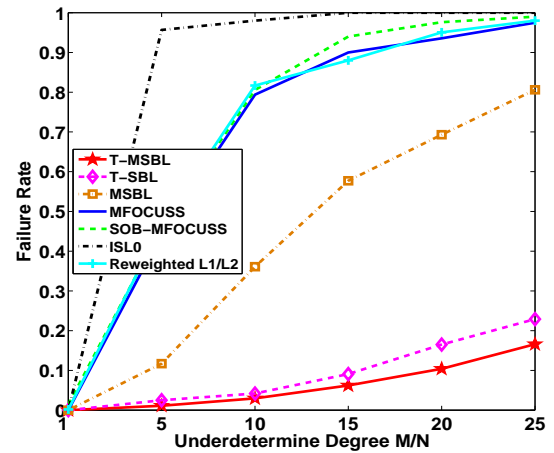}
\caption{Performance comparison in highly underdetermined cases. }
\label{fig:MN}
\end{figure}

Most published works only compared algorithms in mildly underdetermined cases, namely, the ratio of $M/N$ was about $2 \sim 5$. However, in some applications such as neuroimaging, one can easily have $N \approx 100$ and $M \approx 100000$. So, in this experiment we compare the algorithms in the highly underdetermined cases when $N$ was fixed at 25 and $M/N$ varied from 1 to 25. The source number $K$ was 12, and the measurement vector number $L$ was 4. SNR was 25 dB. Different to previous experiments, all the sources were AR(1) processes but with different AR coefficients. Their AR coefficients were uniformly chosen from $(0.5,1)$ at random. Results are presented in Fig.\ref{fig:MN}, from which we can see that when $M/N \geq 10,$ all the compared algorithms had large errors. In contrast, our proposed algorithms  had much lower errors. Note that due to the performance trade-off between $N$ and $M$, if one increases $N$,  algorithms can keep the same recovery performance for larger $M/N$.

\subsection{Recovery Performance for Different Kinds of Sources}
\label{exp:varyARp}

In previous experiments all the sources were AR(1) processes. Although we have pointed out that for small $L$ modeling sources by AR(1) processes is sufficient, here we carry out an experiment to show our algorithms maintaining the same superiority for various time series. Since from previous experiments we have seen that T-SBL has similar performance to T-MSBL,  and that MSBL has the best performance among the compared algorithms, in this experiment we only compare T-MSBL with MSBL.

\begin{figure}[htb]
\begin{minipage}[b]{.48\linewidth}
  \centering
  \centerline{\epsfig{figure=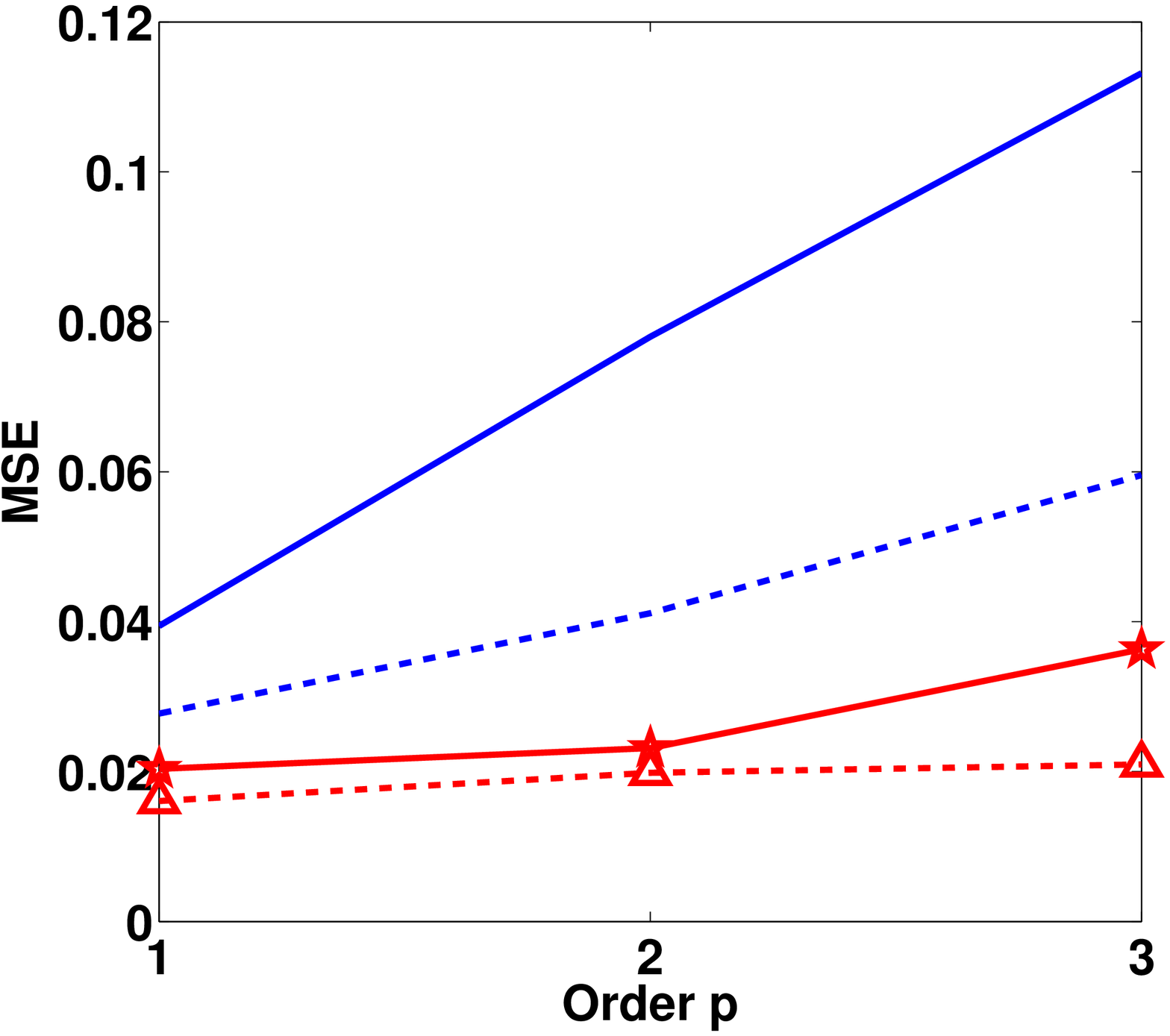,width=4.5cm}}
  \centerline{\footnotesize{(a)}}
\end{minipage}
\hfill
\begin{minipage}[b]{0.48\linewidth}
  \centering
  \centerline{\epsfig{figure=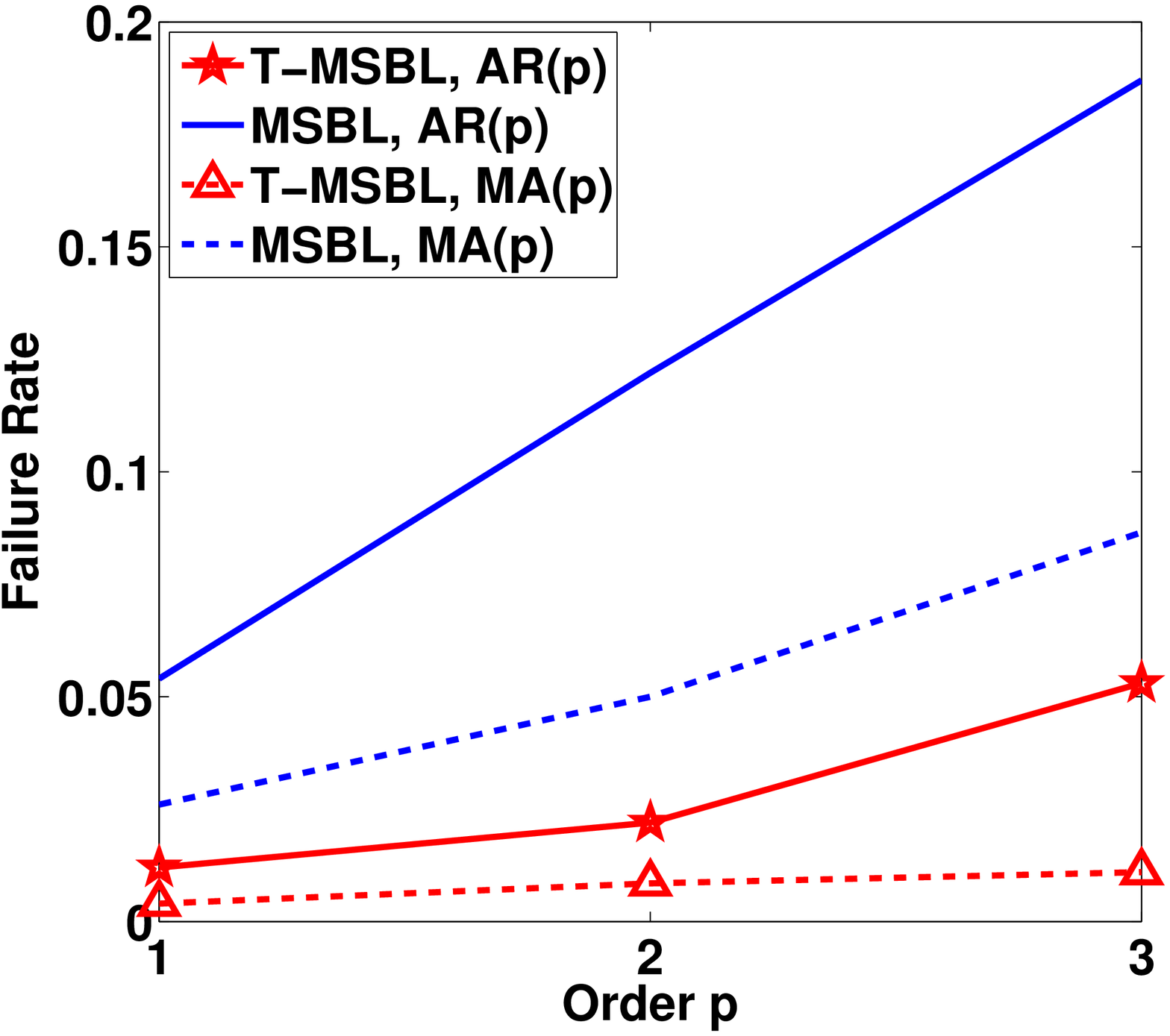,width=4.5cm}}
  \centerline{\footnotesize{(b)}}
\end{minipage}
\caption{Performance of T-MSBL and MSBL for different AR(p) sources and different MA(p) sources measured in terms of MSE and failure rates.}
\label{fig:Simulation_varyARp}
\end{figure}

The dictionary matrix was of the size $25 \times 125$. $L$ was 4. $K$ was 14. SNR was 25dB. First we generated sources as three kinds of AR processes, i.e. $\mathrm{AR}(p)$ ($p=1,2,3$). All the AR coefficients were randomly uniformly chosen from the feasible regions such that the processes were stable. We examined the algorithms' performance as a function of the AR order $p$. Results are given in Fig.\ref{fig:Simulation_varyARp}, showing that T-MSBL again outperformed MSBL. With large $p$, the performance gap between the two algorithms increased. We repeated the previous experiment with the same experiment settings except that we replaced the $\mathrm{AR}(p)$ sources by moving-averaging sources $\mathrm{MA}(p)$ ($p=1,2,3$). The MA coefficients were uniformly chosen from $(0,1]$ at random. Again, we obtained the same results. These results imply that our algorithms maintain their superiority for various temporally structured sources, not only AR processes.

\subsection{Recovery Ability at Different Noise Levels}
\label{exp:varySNR}

\begin{figure}[htb]
\begin{minipage}[b]{.48\linewidth}
  \centering
  \centerline{\epsfig{figure=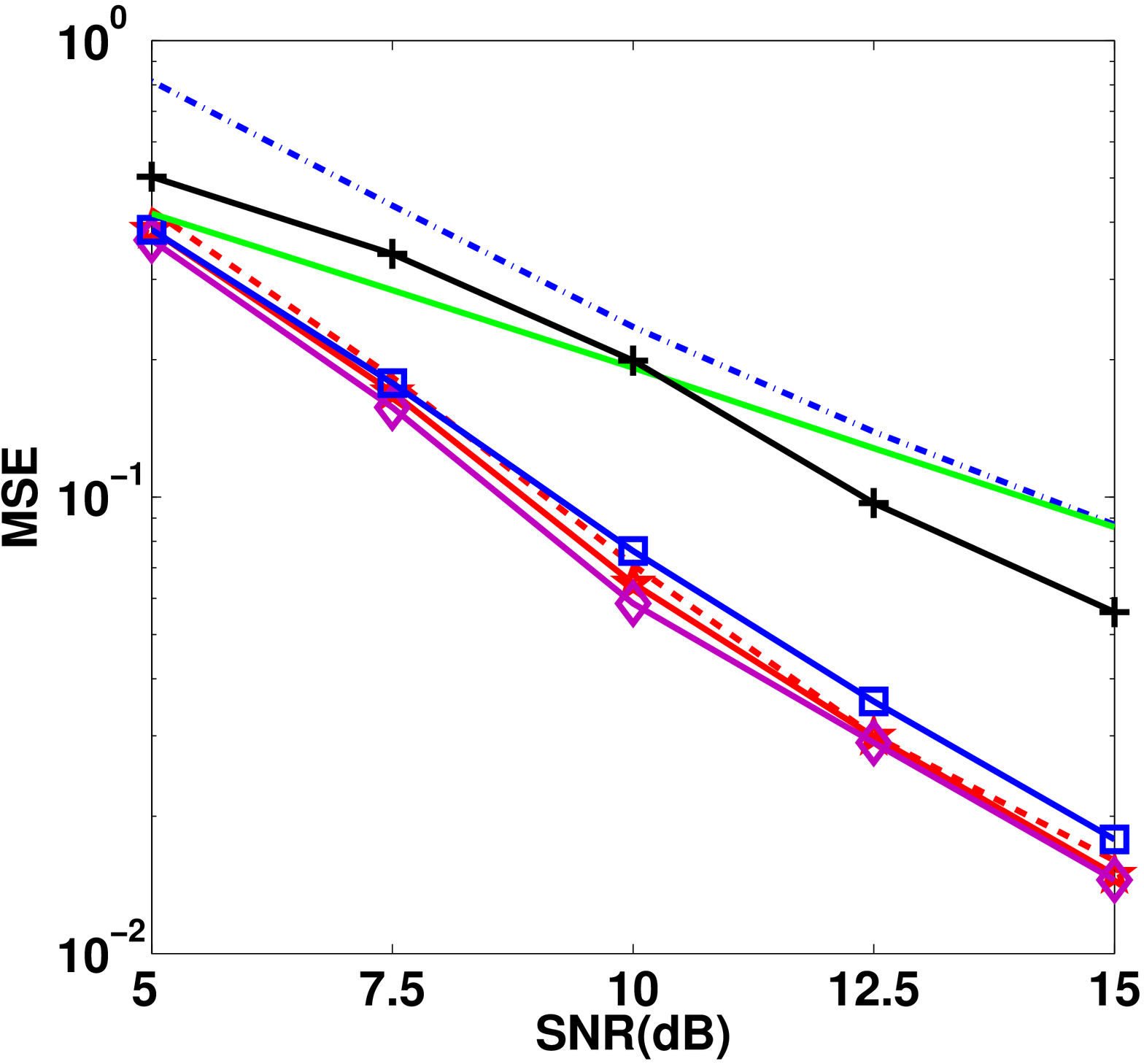,width=4.5cm}}
  \centerline{\footnotesize{(a)}}
\end{minipage}
\hfill
\begin{minipage}[b]{0.48\linewidth}
  \centering
  \centerline{\epsfig{figure=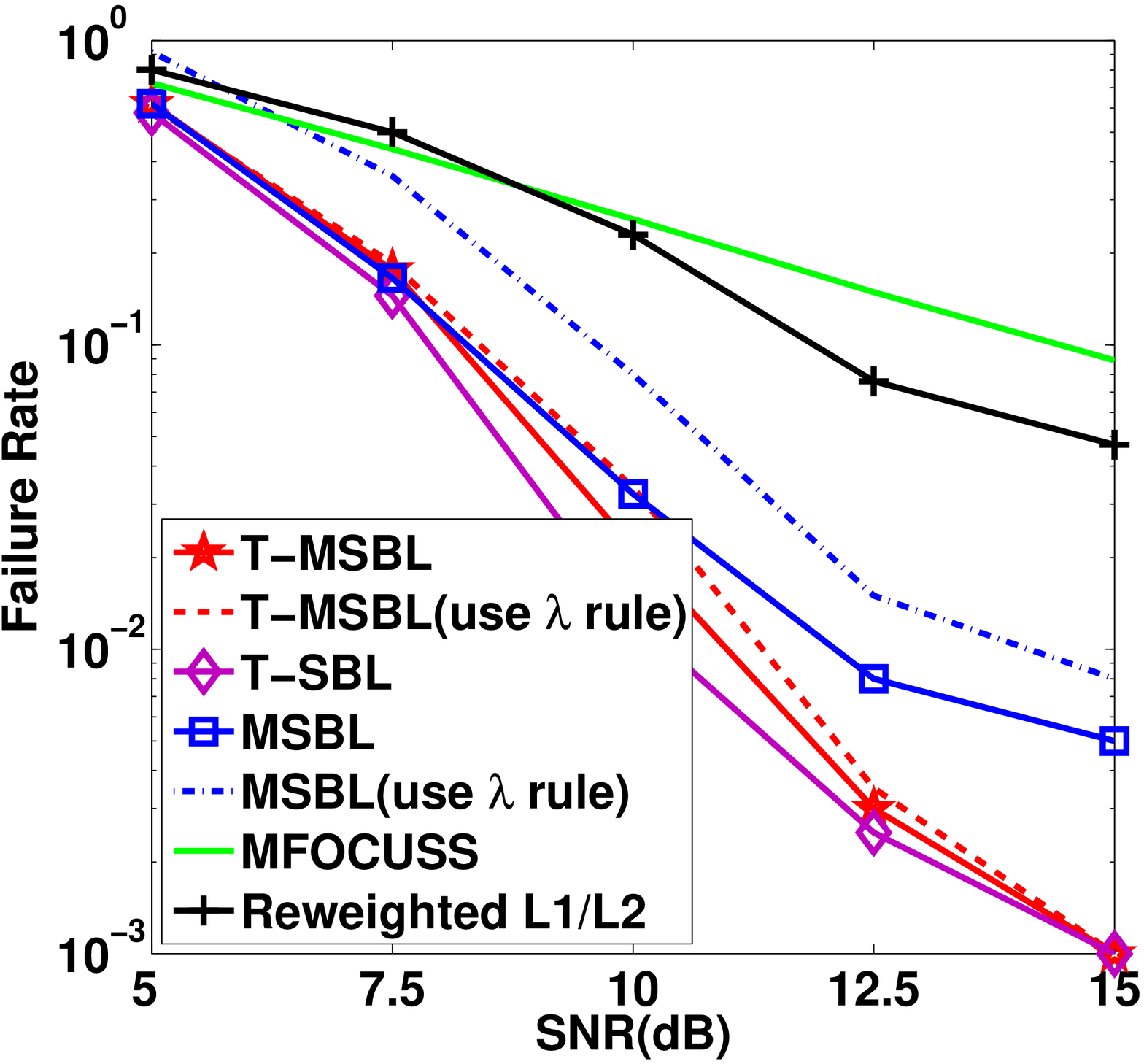,width=4.5cm}}
  \centerline{\footnotesize{(b)}}
\end{minipage}
\caption{Performance of various algorithms at different noise levels.}
\label{fig:Simulation_varySNR}
\end{figure}

From previous experiments we have seen that the proposed algorithms  significantly outperformed all the compared algorithms in noiseless scenarios and mildly noisy cases, even though to derive T-MSBL  we used the approximation (\ref{equ:approximationFormula}) which takes the equal sign only when $\mathbf{B}=\mathbf{I}$ (no temporal correlation) or $\lambda = 0$ (no noise). Some natural questions may be raised: What is the performance of T-SBL and T-MSBL in strongly noisy cases? Is it still beneficial to exploit temporal correlation in these cases? To answer these questions, we carry out the following experiment.

The dictionary matrix was of the size $25 \times 125$. The number of measurement vectors $L$ was 4. The source number $K$ was 7. All the sources were AR(1) processes and the temporal correlation of each source was 0.8. SNR varied from 5 dB to 15 dB. The experiment was repeated 2000 trials. We compared the proposed T-SBL, T-MSBL with three representative algorithms, i.e. MSBL, MFOCUSS, and Reweighted $\ell_1/\ell_2$.

Note that in low SNR cases, the estimated $\mathbf{B}$ of T-SBL and T-MSBL can include large errors, and thus the estimated amplitudes of sources are  distorted. To reduce the distortion, we set $\mathbf{B}=\mathbf{I}$ once the number of nonzero $\gamma_i$ was less than $N$ during the learning procedure. The reason is that the role of $\mathbf{B}$ is to prevent T-SBL/T-MSBL from arriving at local minima; once the algorithms approach global minima very closely, $\mathbf{B}$ is no longer useful.

Also note that the $\lambda$ learning rules of T-SBL, T-MSBL and MSBL may not lead to optimal performance in low SNR cases. To avoid the potential disturbance of these $\lambda$ learning rules, we provided the three SBL algorithms with the optimal $\lambda^*$'s, which were obtained by the exhaustive search method stated previously.

Figure \ref{fig:Simulation_varySNR} shows that T-SBL and T-MSBL outperformed other algorithms in all the noise levels. This implies that even in low SNR cases exploiting temporal correlation of sources is beneficial.

But we want to emphasize that although the $\lambda$ learning rules of the three SBL algorithms may not be optimal in low SNR cases, our proposed $\lambda$ learning rules can lead to near-optimal performance, compared to the one of MSBL. To see this, we ran T-MSBL and MSBL again, but this time both algorithms used their $\lambda$ learning rules. T-MSBL used the modified version of the $\lambda$ learning rule (\ref{equ:lambdaRule_simplified}), i.e. setting the off-diagonal elements of $\mathbf{\Phi} \mathbf{\Gamma} \mathbf{\Phi}^T$ to zeros. The results (Fig. \ref{fig:Simulation_varySNR}) show that MSBL had very poor performance when using its $\lambda$ learning rule. In contrast, T-MSBL's performance was very close to its performance when using its optimal $\lambda^*$ \footnote{T-SBL had the same behavior. But for clarity we do not present its performance curve.}. The results indicate our proposed algorithms are advantageous in practical applications, since in practice the optimal $\lambda^*$'s are difficult to obtain, if not impossible.

\subsection{Temporal Correlation: Beneficial or Detrimental?}
\label{exp:goodbadofTC}

From previous experiments one may think that temporal correlation is always harmful to algorithms' performance, at least not helpful. However, in this experiment we will show that when SNR is high, the performance of our proposed algorithms increases with increasing temporal correlation.

\begin{figure}
\centering
\includegraphics[width=8cm,height=6.5cm]{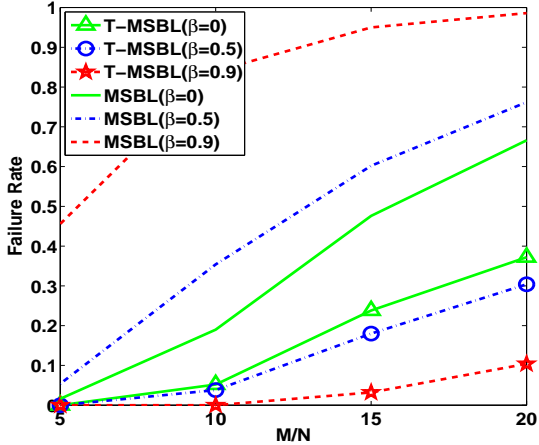}
\caption{Behaviors of MSBL and T-MSBL  at different temporal correlation levels when $\mathrm{SNR}=50\mathrm{dB}$.}
\label{fig:TCbenefit}
\end{figure}

We set $N=25$, $L=4$, $K=14$, and $\mathrm{SNR} = 50 \mathrm{dB}$. The underdeterminacy ratio $M/N$ varied from 5 to 20. Sources were generated as AR(1) processes with the common AR coefficient $\beta$. We considered the performance of T-MSBL and MSBL in three cases, i.e. the temporal correlation $\beta$ was 0, 0.5, and 0.9, respectively. Results are shown in Fig.\ref{fig:TCbenefit}. As expected, the performance of MSBL deteriorated with increasing temporal correlation. But the behavior of T-MSBL was rather counterintuitive. It is surprising that the best performance of T-MSBL was not achieved at $\beta=0$, but at $\beta=0.9$. Clearly, high temporal correlation enabled T-MSBL to handle more highly underdetermined problems. For example, its performance at $M/N=20$ with $\beta=0.9$ was better than that at $M/N=15$ with $\beta=0.5$ or $\beta=0$. The same phenomenon was observed in  noiseless cases as well, and was observed for T-SBL.

The results indicating that temporal correlation is helpful may appear
counterintuitive at first glance. A closer examination of the sparse recovery problems indicates a plausible explanation. There are two elements to the sparse recovery task; one is the location of the nonzero entries and the other is the value for the nonzero entries. Both tasks interact and combine to determine the overall performance. Correlation helps the estimation of the values for the nonzero entries and this may be important for the problem when dealing with finite matrices and may be lost when dealing with limiting results as the matrix dimension go to infinity. A more rigorous study of the interplay between estimation of the values and estimation of the locations is an interesting topic.

\subsection{An Extreme Experiment on the Importance of Exploiting Temporal Correlation}
\label{exp:extremeEg}

It may be natural to take for granted that in noiseless cases, when source vectors are almost identical, algorithms have almost the same performance as in the case when only one measurement vector is available. In the following we show that it is not the case.

We designed a noiseless experiment. First, we generated a Hadamard matrix of the size $128 \times 128$. From the matrix, 40 rows were randomly selected in each trial and formed a dictionary matrix of the size $40 \times 128$. The source number $K$ was 12, and the measurement vector number $L$ was 3. Sources were generated as AR(1) processes with the common AR coefficient $\beta$, where $\beta = \mathrm{sign}(C)(1-10^{-|C|})$. We varied $C$ from -10 to 10 in order to see how algorithms behaved  when the absolute temporal correlation, $|\beta|$, approximated to 1.

\begin{figure}[htb]
\begin{minipage}[b]{.48\linewidth}
  \centering
  \centerline{\epsfig{figure=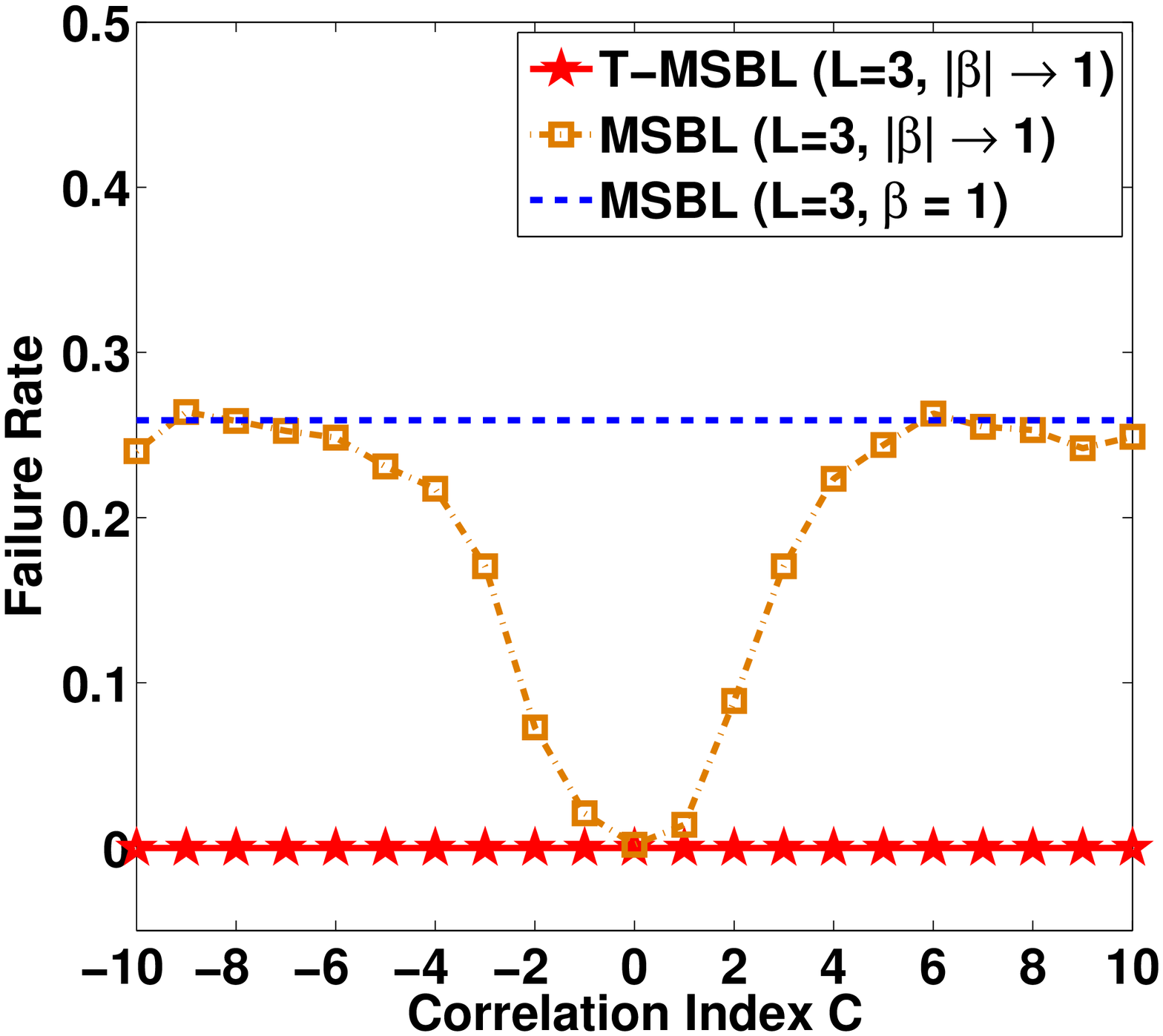,width=4.5cm}}
  \centerline{\footnotesize{(a)}}
\end{minipage}
\hfill
\begin{minipage}[b]{0.48\linewidth}
  \centering
  \centerline{\epsfig{figure=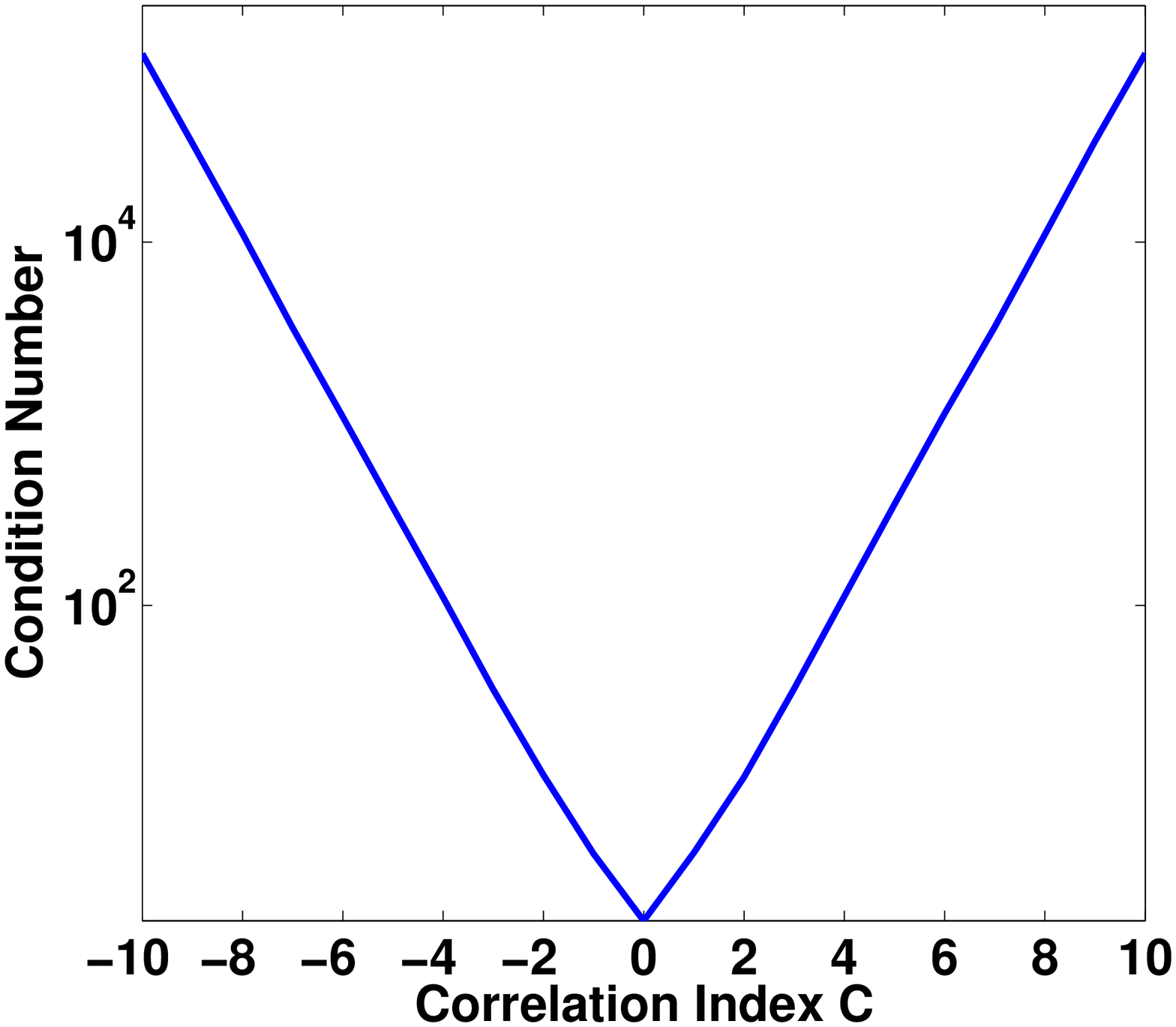,width=4.5cm}}
  \centerline{\footnotesize{(b)}}
\end{minipage}
\caption{(a) The performance and (b) the condition numbers of the submatrix formed by sources when the temporal correlation approximated to 1. The temporal correlation  $\beta=\mathrm{sign}(\mathrm{C})(1-10^{-|\mathrm{C}|})$, where $\mathrm{C}$ was the correlation index varying from -10 to 10. }
\label{fig:extremeCase}
\end{figure}

Figure \ref{fig:extremeCase} (a) shows the performance curves of T-MSBL and MSBL when $|\beta|\rightarrow 1$, and also shows the performance curve of MSBL when $\beta = 1$.  We observe an interesting phenomenon. First, as $|\beta|\rightarrow 1$, MSBL's performance closely approximated to its performance in the case of $\beta=1$. It seems to make sense, because when $|\beta| \rightarrow 1$, every source vector provides almost the same information on locations and amplitudes of nonzero elements. Counter-intuitively, no matter how close $|\beta|$ was to 1, the performance of T-MSBL did not change. Figure \ref{fig:extremeCase} (b) shows the averaged condition numbers of the submatrix formed by the sources (i.e. nonzero rows in $\mathbf{X}_{\mathrm{gen}}$) at different correlation levels. We can see that the condition numbers increased with the increasing temporal correlation. This suggests that T-MSBL was not sensitive to the ill-condition issue in the source matrix, while MSBL is very sensitive. Although not shown here, we found that T-SBL had the same behavior as T-MSBL, while other MMV algorithms had the same behaviors as MSBL. The phenomenon was also observed when using other dictionary matrices, such as random Gaussian matrices.

These results emphasize the importance of exploiting the temporal correlation, and also motivate future theoretical studies on the temporal correlation and the ill-condition issue of source matrices.

\section{Discussions}
\label{sec:discussion}

Although there are a few works trying to exploit temporal correlation in the MMV model, based on our knowledge no works have explicitly studied the effects of temporal correlation, and no existing algorithms are  effective in the presence of such correlation. Our work is a starting point in the direction of considering temporal correlation in the MMV model. However, there are many issues that are unclear so far. In this section we discuss some of them.

\subsection{The Matrix $\mathbf{B}$: Trade-off Between Accurately Modeling and Preventing Overfitting}

In our algorithm development we used one single matrix $\mathbf{B}$ as the covariance matrix (up to a scalar) for each source  model in order to avoid overfitting. Mathematically, it is straightforward to extend our algorithms to use multiple  matrices to capture the covariance structures of sources. For example, one can classify sources into several groups, say $G$ groups,  and the sources in a group are all assigned by a common  matrix $\mathbf{B}_i$ ($i=1,\cdots,G,\; G \ll M$) as the covariance matrix (up to a scalar). It seems that this extension can better capture the covariance structures of sources while still avoiding overfitting. However, we find that this extension (even for $G=2$) has much poorer performance than our proposed algorithms and MSBL. One possible reason is that during the early stage of the learning procedure of our algorithms, the estimated sources from each iteration are far from the true sources, and thus grouping them based on their covariance structures is difficult, if not impossible. The grouping error may cause avalanche effect, leading to the noted poor performance. Reducing the grouping error and more accurately capturing the temporal correlation structures is an area for future work.

However, as we have stated, $\mathbf{B}$ plays a role of whitening each source. In our recent work \cite{Zhilin_ICASSP2011,Zhilin_ICML2011} we found that the operation $\mathbf{X}_{i\cdot} \mathbf{B}^{-1} \mathbf{X}_{i\cdot}^T$ ($\forall i$) can replace the row-norms (such as the $\ell_2$ norm and the $\ell_{\infty}$ norm) in iterative reweighted $\ell_2$ and $\ell_1$ algorithms for the MMV model, functioning as a row regularization. This indicates that using one single matrix $\mathbf{B}$ may be a better method than using multiple matrices $\mathbf{B}_1,\cdots,\mathbf{B}_G$.

On the other hand, there may be many ways to parameterize and estimate $\mathbf{B}$. In this work we provide a general method to estimate $\mathbf{B}$. In \cite{Zhilin2010} we proposed a method to parameterize $\mathbf{B}$ by a hyperparameter $\beta$, i.e.,
\begin{equation}
\mathbf{B} =  \left[
\begin {array}{cccc}
1       & \beta   &  \cdots   & \beta^{L-1} \\
\beta   & 1       &  \cdots   & \beta^{L-2} \\
\vdots  & \vdots  &   \ddots   & \vdots     \\
\beta^{L-1} & \beta^{L-2}     &  \cdots   & 1
\end {array}
\right] \nonumber
\end{equation}
which equivalently assumes the sources are AR(1) processes with the common AR coefficient $\beta$. The resulting algorithms have good performance as well. Also, for low SNR cases in our experiments, we added an identity matrix (with a scalar) to the estimated $\mathbf{B}$ in T-MSBL, and achieved satisfying performance. All these imply that $\mathbf{B}$ could have many forms. Finding the forms that are advantageous in strongly noisy environments is an important issue and needs further study.

\subsection{The Parameter $\lambda$: Noise Variance or Regularization Parameter?}

In our algorithms the covariance matrix of the multi-channel noise $\mathbf{V}_{\cdot i}$ ($i=1,\cdots,L$) is $\lambda \mathbf{I}_N$ with the implicit assumption that each channel noise has the same variance $\lambda$. It is straightforward to extend our algorithms to consider the general noise covariance matrix $\mathrm{diag}([\lambda_1,\cdots,\lambda_N])$, i.e. assuming different channel noise have different variance. However, this largely increases parameters to estimate, and thus we may once again encounter an overfitting problem (similar to the overfitting problem in learning the matrix $\mathbf{B}_i$).

Some works \cite{David2010NI,Qiu2010} considered  alternative noise covariance models. In \cite{Qiu2010} the authors assumed that the covariance matrix of multi-channel noise is $\lambda \mathbf{C}$, instead of $\lambda \mathbf{I}_N$, where $\mathbf{C}$ is a known positive definite and symmetric matrix and $\lambda$ is an unknown noise-variance parameter. This model may better capture the noise covariance structures, but generally one does not know the exact value of  $\mathbf{C}$. Thus there is no clear  benefit  from this covariance model. In \cite{David2010NI}, instead of deriving a learning rule for the noise covariance inside the SBL framework, the authors estimated the noise covariance by a method independent of the SBL framework. But this method is based on a large number of measurement vectors, and has a high computational load.

On the other hand, due to the works in \cite{David2010latent,David2010reweighting,David2008NIPS}, which connected SBL algorithms to traditional convex relaxation methods such as Lasso \cite{Lasso} and Basis Pursuit Denoising \cite{BP}, it was  found that $\lambda$ is functionally the same as the regularization parameters of those convex relaxation algorithms. This suggests the use of methods such as the modified L-curve procedure \cite{Rao2003} or the cross-validation  \cite{Lasso,BP} to choose  $\lambda$ especially in strongly noisy environments. It is also interesting to see that SBL algorithms could adopt the continuation strategies \cite{NESTA,Hale2007}, used in Lasso-type algorithms, to adjust the value of $\lambda$ for better recovery performance or faster speed.

However, if some channels contain very large noise (e.g. outliers) and the number of such channels is very small, then as suggested in \cite{Wright2009}, we can extend the dictionary matrix $\mathbf{\Phi}$ to $[\mathbf{\Phi},\mathbf{I}]$ and perform any sparse signal recovery algorithms without modification. The estimated `sources' associated with the identity dictionary matrix are these large noise components.

\subsection{Connections to Other Models}

In fact, our bSBL framework is a block sparsity model \cite{Negahban2009,groupLasso,Eldar2009}, and thus the derived T-SBL algorithm can be directly used for this model. Compared to most existing algorithms derived in this model \cite{Negahban2009,groupLasso,Eldar2010TSP}, an important difference is that T-SBL considers the correlation within each block.

The time-varying sparsity model \cite{Vaswani2008,Philip2010} is another related model. Different to our MMV model that assumes the support of each source vector is the same, the time-varying sparsity model assumes the support is slowly time-varying. It is interesting to note that this model can be approximated by concatenation of several MMV models, where in each MMV model the support does not change. Thus our proposed T-SBL and T-MSBL can be used for this model. The results are appealing, as shown in our recent work \cite{Zhilin_ICML2011}.

It should be noted that the proposed algorithms can be directly used for the SMV model. In this case the matrix $\mathbf{B}$ reduces to a scalar, and the $\gamma_i$ learning rules are the same as the one in the basic SBL algorithm \cite{David2004IEEE}. But due to the effective $\lambda$ learning rules, our algorithms are superior to the basic SBL algorithm, especially in noisy cases.

\section{Conclusions}

We addressed a multiple measurement vector (MMV) model in practical scenarios, where the source vectors are temporally correlated and the number of measurement vectors is small due to the common sparsity constraint. We showed that existing algorithms have poor performance when  temporal correlation is present, and thus they have limited ability in practice. To solve this problem, we proposed a block sparse Bayesian learning framework, which allows for easily modeling the temporal correlation and incorporating this information into derived algorithms. Based on this framework, we derived two algorithms, namely, T-SBL and T-MSBL. The latter can be seen as an extension of MSBL by replacing the $\ell_2$ norm imposed on each source with a Mahalanobis distance measure. Extensive experiments have shown that the proposed algorithms have superior performance to many state-of-the-art algorithms. Theoretical analysis also has shown that the proposed algorithms have desirable global and local minimum properties.

\section*{Acknowledgement}
Z.Z would like to thank  Dr. David Wipf for his considerable help with the study of SBL, Ms. Jing Wan for kind help in performing some experiments, Mr. Tim Mullen for kind help in the paper writing, Dr. Rafal Zdunek for providing the code of SOB-MFOCUSS, and Mr. Md Mashud Hyder for providing the code of ISL0. The authors thank the reviewers for their helpful comments and especially thank a reviewer for the idea of using multiple covariance matrices, which is discussed in Section VII.A.

\section*{Appendix}
\subsection{Outline of the Proof of Theorem \ref{theorem:global}}

Since the proof is a generalization of the Theorem 1 in \cite{David2010NI}, we only give an outline.

For convenience we consider the equivalent model (\ref{equ:blocksparsemodel}). Let $\widehat{\mathbf{x}}$ be computed using $\widehat{\mathbf{x}} = (\lambda {\widehat{\mathbf{\Sigma}}_0}^{-1} + \mathbf{D}^T \mathbf{D})^{-1} \mathbf{D}^T \mathbf{y}$ with $\widehat{\mathbf{\Sigma}}_0 = \mathrm{diag}\{\widehat{\gamma}_1 \widehat{\mathbf{B}}_1,\cdots, \widehat{\gamma}_M \widehat{\mathbf{B}}_M \}$, and $\widehat{\boldsymbol{\gamma}} \triangleq [\widehat{\gamma}_1,\cdots,\widehat{\gamma}_M]$  is obtained by globally minimizing the cost function for given $\widehat{\mathbf{B}}_i \,(\forall i)$ \footnote{In the proof we fix $\widehat{\mathbf{B}}_i$ because we will see $\widehat{\mathbf{B}}_i$ has no effect on the global minimum property.}:
\begin{eqnarray}
\mathcal{L}(\gamma_i) = \mathbf{y}^T \mathbf{\Sigma}_y^{-1} \mathbf{y} + \log|\mathbf{\Sigma}_y|. \nonumber
\end{eqnarray}
It can be shown \cite{David2010NI} that when $\lambda \rightarrow 0$ (noiseless case), the above problem is equivalent to
\begin{eqnarray}
\min: &\quad& g(\mathbf{x}) \triangleq\min_{\mathbf{\boldsymbol{\gamma}}} \big[ \mathbf{x}^T \mathbf{\Sigma}_0^{-1} \mathbf{x}  +  \log|\mathbf{\Sigma}_y|     \big]   \label{equ:g}\\
\mathrm{s.t.}: &\quad &  \mathbf{y} = \mathbf{Dx}
\end{eqnarray}
So we only need to show the global minimizer of (\ref{equ:g}) satisfies the property stated in the theorem.

Assume in the noiseless problem $\mathbf{Y}=\mathbf{\Phi} \mathbf{X}$, $\mathbf{\Phi}$ satisfies the URP condition \cite{Gorodnitsky1997}. For its any solution $\widehat{\mathbf{X}}$, denote the number of nonzero rows by $K$. Thus following the method in \cite{David2010NI}, we can show the above $g(\mathbf{x})$ satisfies
\begin{eqnarray}
g(\mathbf{x}) = \mathcal{O}(1) + \big(NL - \min[NL, KL  ]\big)\log \lambda,
\label{equ:g_final}
\end{eqnarray}
providing $\widehat{\mathbf{B}}_i$ is full rank. Here we adopt the notation $f(s) = \mathcal{O}(1)$ to indicate that $|f(s)| < C_1$ for all $s < C_2$, with $C_1$ and $C_2$ constants independent of $s$. Therefore, by globally minimizing (\ref{equ:g_final}), i.e. globally minimizing (\ref{equ:g}), $K$ will achieve its minimum value, which will be shown to be $K_0$, the number of nonzero rows in $\mathbf{X}_{\mathrm{gen}}$.

According to the result in \cite{Cotter2005,Donoho2003}, if $\mathbf{X}_{\mathrm{gen}}$ satisfies
\begin{eqnarray}
K_0  <  \frac{N + L}{2}  \nonumber
\end{eqnarray}
then there is no other solution (with $K$ nonzero rows) such that $\mathbf{Y}=\mathbf{\Phi}\mathbf{X}$ with $K  <  \frac{N + L}{2}$. So, $K \geq K_0$, i.e. the minimum value of $K$ is $K_0$. Once $K$ achieves its minimum, we have $\widehat{\mathbf{X}} = \mathbf{X}_{\mathrm{gen}}$.

In summary, the global minimum solution $\widehat{\boldsymbol{\gamma}}$ leads to the  solution  that equals to the unique sparsest solution $\mathbf{X}_{\mathrm{gen}}$. And we can see, providing $\widehat{\mathbf{B}}_i$ is full rank, it does not affect the conclusion.

\subsection{Proof of Lemma \ref{lemma:b_AGamma}}

Re-write the equation $\mathbf{y}^T \mathbf{\Sigma}_y^{-1} \mathbf{y} = C$ by $\mathbf{y}^T \mathbf{u} = C$, where $\mathbf{u} \triangleq \mathbf{\Sigma}_y^{-1} \mathbf{y} = \big(\lambda \mathbf{I} + \mathbf{D} \mathbf{\Sigma}_0 \mathbf{D}^T \big)^{-1} \mathbf{y}$, from which we have $\mathbf{y} - \lambda \mathbf{u} = \mathbf{D} \mathbf{\Sigma}_0 \mathbf{D}^T \mathbf{u} = \mathbf{D} (\mathbf{\Gamma} \otimes \mathbf{B}) \mathbf{D}^T \mathbf{u} = \mathbf{D} (\mathbf{I}_M \otimes \mathbf{B})(\mathbf{\Gamma} \otimes \mathbf{I}_L) \mathbf{D}^T \mathbf{u} = \mathbf{D} (\mathbf{I}_M \otimes \mathbf{B}) \mathrm{diag}(\mathbf{D}^T \mathbf{u}) \mathrm{diag}(\mathbf{\Gamma} \otimes \mathbf{I}_L) = (\mathbf{\Phi} \otimes \mathbf{B}) \mathrm{diag}(\mathbf{D}^T \mathbf{u}) (\mathbf{\boldsymbol{\gamma}} \otimes \mathbf{1}_L)$. It can be seen that the  matrix $\mathbf{A} \triangleq (\mathbf{\Phi} \otimes \mathbf{B}) \mathrm{diag}(\mathbf{D}^T \mathbf{u})$ is full row rank.

\subsection{Proof of Theorem \ref{theorem:local}}

The proof follows along the lines of Theorem 2 in \cite{David2004IEEE} using our Lemma \ref{lemma:concavity} and Lemma \ref{lemma:b_AGamma}. Consider the optimization problem:
\begin{equation} \label{equ:newProblem}
\left\{
\begin{array}{lll}
\min: &\quad& f(\mathbf{\boldsymbol{\gamma}}) \triangleq  \log|\lambda \mathbf{I} + \mathbf{D} \mathbf{\Sigma}_0 \mathbf{D}^T| \\
\mathrm{s.t.}: &\quad& \mathbf{A} \cdot (\mathbf{\boldsymbol{\gamma}} \otimes \mathbf{1}_L) = \mathbf{b}  \\
 &\quad& \boldsymbol{\gamma} \succeq \mathbf{0}
\end{array}
\right.
\end{equation}
where $\mathbf{A}$ and $\mathbf{b}$ are defined in Lemma \ref{lemma:b_AGamma}. From Lemma \ref{lemma:concavity} and Lemma \ref{lemma:b_AGamma} we can see the optimization problem (\ref{equ:newProblem}) is optimizing a concave function over a closed, bounded convex polytope. Obviously, any local minimum of $\mathcal{L}$, e.g. $\boldsymbol{\gamma}^*$, must also be a local minimum of the above optimization problem with $C=\mathbf{y}^T \big(\lambda \mathbf{I} + \mathbf{D} (\mathbf{\Gamma}^* \otimes \mathbf{B}) \mathbf{D}^T  \big)^{-1} \mathbf{y}$, where $\mathbf{\Gamma}^* \triangleq \mathrm{diag}(\boldsymbol{\gamma}^*)$. Based on the Theorem 6.5.3 in \cite{LuenbergerBook} the minimum of (\ref{equ:newProblem}) is achieved at an extreme point. Further, based on the Theorem in Chapter 2.5 of \cite{LuenbergerBook} the extreme point is a BFS to
\begin{equation}
\left\{
\begin{array}{ll}
\mathbf{A} \cdot (\mathbf{\boldsymbol{\gamma}} \otimes \mathbf{1}_L) = \mathbf{b}  \\
\boldsymbol{\gamma} \succeq \mathbf{0}
\end{array}
\right. \nonumber
\end{equation}
which indicates $\|\boldsymbol{\gamma}\|_0 \leq NL$.

\subsection{Proof of Lemma \ref{lemma:localSolution}}

For convenience we first consider the case of $K = N$. Let $\widetilde{\boldsymbol{\gamma}}$ be the vector consisting of nonzero elements in $\widehat{\boldsymbol{\gamma}}$, and $\widetilde{\mathbf{\Phi}}$ be a matrix consisting of the columns of $\mathbf{\Phi}$ whose indexes are the same as those of nonzero elements in $\widehat{\boldsymbol{\gamma}}$. Thus, the equation $\mathbf{Y}=\mathbf{\Phi} \widehat{\mathbf{X}}$ can be rewritten as $\mathbf{Y}= \widetilde{\mathbf{\Phi}} \widetilde{\mathbf{X}}$. By transferring it to its equivalent block sparse Bayesian learning model, we have $\mathbf{y}= \widetilde{\mathbf{D}} \widetilde{\mathbf{x}}$, where $\mathbf{y} \triangleq \mathrm{vec}(\mathbf{Y}^T)$, $\widetilde{\mathbf{D}} \triangleq \widetilde{\mathbf{\Phi}} \otimes \mathbf{I}_L$, and $\widetilde{\mathbf{x}} \triangleq \mathrm{vec}(\widetilde{\mathbf{X}}^T)$. Since $\widetilde{\mathbf{D}}$ is a square matrix with full rank, we have $\widetilde{\mathbf{x}} = \widetilde{\mathbf{D}}^{-1} \mathbf{y}$. For convenience, let $\widetilde{\mathbf{x}}_i \triangleq  \widetilde{\mathbf{x}}_{[(i-1)L+1 : iL]}$, i.e. $\widetilde{\mathbf{x}}_i$ consists of elements of $\widetilde{\mathbf{x}}$ with indexes from $(i-1)L+1$ to $iL$. Now consider the cost function $\mathcal{L}$, which becomes
\begin{eqnarray}
\mathcal{L}(\boldsymbol{\gamma})
&=& \sum_{i=1}^N \Big(\frac{\widetilde{\mathbf{x}}_i^T {\mathbf{B}}^{-1} \widetilde{\mathbf{x}}_i}{\widetilde{\gamma}_i} +  L \log \widetilde{\gamma}_i \Big) + M\log|\mathbf{B}|  \nonumber \\
&& + 2\log|\widetilde{\mathbf{D}}|.   \nonumber
\end{eqnarray}
Letting $\frac{\partial \mathcal{L}(\boldsymbol{\gamma})}{\partial \widetilde{\gamma}_i} = 0$ gives
\begin{eqnarray}
\widetilde{\gamma}_i = \frac{1}{L}\widetilde{\mathbf{x}}_i^T {\mathbf{B}}^{-1} \widetilde{\mathbf{x}}_i, \quad i=1,\cdots,K \nonumber
\end{eqnarray}
The second derivative of $\mathcal{L}$ at $\widetilde{\gamma}_i = \frac{1}{L}\widetilde{\mathbf{x}}_i^T {\mathbf{B}}^{-1} \widetilde{\mathbf{x}}_i$ is given by
\begin{eqnarray}
\frac{\partial^2 \mathcal{L}(\boldsymbol{\gamma})}{\partial {\widetilde{\gamma}_i}^2} \Big|_{\widetilde{\gamma}_i = \widetilde{\mathbf{x}}_i^T {\mathbf{B}}^{-1} \widetilde{\mathbf{x}}_i}  = \frac{\widetilde{\mathbf{x}}_i^T {\mathbf{B}}^{-1} \widetilde{\mathbf{x}}_i}{\widetilde{\gamma}_i^3}. \nonumber
\end{eqnarray}
Since $\mathbf{B}$ is positive definite and $\widetilde{\mathbf{x}}_i \neq \mathbf{0}$, $\frac{\widetilde{\mathbf{x}}_i^T {\mathbf{B}}^{-1} \widetilde{\mathbf{x}}_i}{\widetilde{\gamma}_i^3} > 0$. So $\widetilde{\gamma}_i = \frac{1}{L} \widetilde{\mathbf{x}}_i^T {\widehat{\mathbf{B}}}^{-1} \widetilde{\mathbf{x}}_i \;(i=1,\cdots,K)$ is a local minimum.

If $\|\widehat{\boldsymbol{\gamma}}\|_0 \triangleq K < N$, which implies there exists $\widetilde{\mathbf{x}} \in \mathbb{R}^{KL \times 1}$ such that $\mathbf{y} = \widetilde{\mathbf{D}} \widetilde{\mathbf{x}}$, then we can expand the matrix $\widetilde{\mathbf{D}}$ to a full-rank square matrix $[\widetilde{\mathbf{D}},\mathbf{D}_e]$ by adding an arbitrary full column-rank matrix $\mathbf{D}_e$. And we expand $\widetilde{\mathbf{x}}$ to $[\widetilde{\mathbf{x}}^T, \boldsymbol{\varepsilon}^T]^T$, where $\boldsymbol{\varepsilon} \in \mathbb{R}^{(N-K)L \times 1}$ and $\boldsymbol{\varepsilon} \rightarrow \mathbf{0}$. Therefore, $[\widetilde{\mathbf{D}},\mathbf{D}_e] [\widetilde{\mathbf{x}}^T, \boldsymbol{\varepsilon}^T]^T \rightarrow \widetilde{\mathbf{D}} \widetilde{\mathbf{x}} = \mathbf{y}$. Similarly, we also expand $\widetilde{\boldsymbol{\gamma}}$ to $[\widetilde{\boldsymbol{\gamma}}^T, \boldsymbol{\zeta}^T]^T$ with $\boldsymbol{\zeta} \rightarrow \mathbf{0}$. Then, following the above steps, we can obtain the same result. Therefore, we finish the proof.

\bibliographystyle{IEEEtran}

\bibliography{IEEEabrv,sparsebibfile}

\begin{biography}[{\includegraphics[width=1.0in,height=1.25in,clip,keepaspectratio]{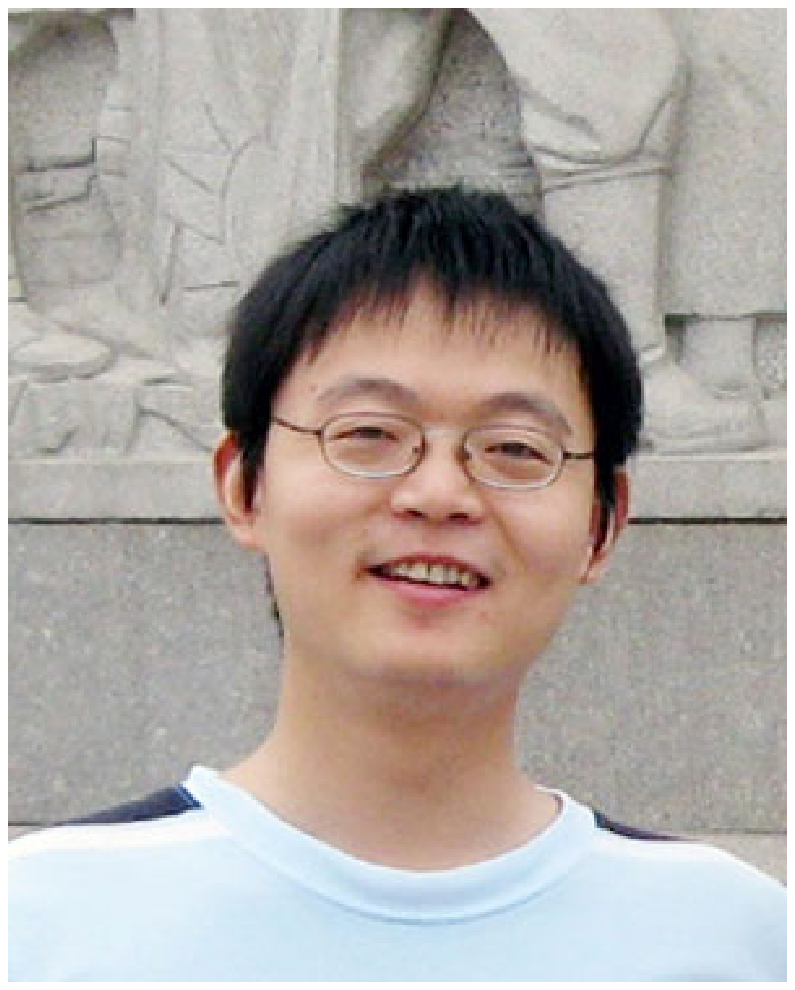}}]{Zhilin Zhang} (S'08) received the B.S. degree in automatics and the M.S. degree in electrical engineering from the University of Electronic Science and Technology of China. Since 2007 he has been working toward the Ph.D. degree in the Department of Electrical and Computer Engineering at University of California, San Diego.

His research interests include sparse signal recovery/compressed sensing, blind source separation, neuroimaging, computational and cognitive neuroscience.

\end{biography}

\begin{biography}[{\includegraphics[width=1.0in,height=1.25in,clip,keepaspectratio]{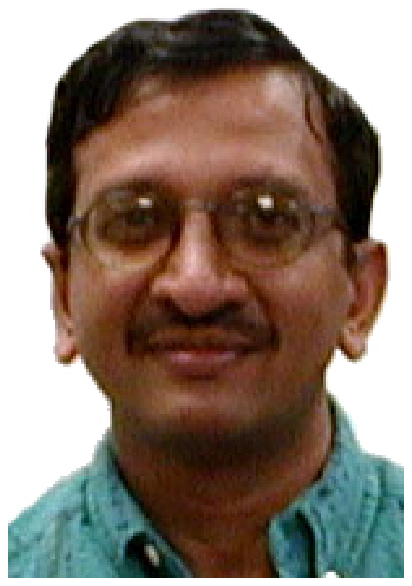}}]{Bhaskar D. Rao} (F'00) received the B.Tech. degree in electronics and
electrical communication engineering from the Indian Institute of
Technology, Kharagpur, India, in 1979 and the M.S. and Ph.D. degrees
from the University of Southern California, Los Angeles, in 1981 and
1983, respectively. Since 1983, he has been with the University of
California at San Diego, La Jolla,
where he is currently a Professor with the Electrical and Computer
Engineering Department and holder of the Ericsson endowed chair in
wireless access networks. His interests are in the areas of digital
signal processing, estimation theory, and optimization theory, with
applications to digital communications, speech signal processing, and
human-computer interactions.

He is the holder of the Ericsson endowed chair in Wireless Access
Networks and is the Director of the Center for Wireless
Communications. His research group has received several paper awards.
His paper received the best paper award at the 2000 speech coding
workshop and his students have received student paper awards at both
the 2005 and 2006 International conference on Acoustics, Speech and
Signal Processing conference as well as the best student paper award
at NIPS 2006. A paper he co-authored  with B. Song and  R. Cruz
received the 2008 Stephen O. Rice Prize Paper Award in the Field of
Communications Systems.  He was elected to the fellow grade in 2000
for his contributions in high resolution spectral estimation. Dr. Rao
has been a member of the Statistical Signal and Array Processing
technical committee, the Signal Processing Theory and Methods
technical committee, the Communications technical committee of the
IEEE Signal Processing Society. He has also served on the editorial
board of the EURASIP Signal Processing Journal.
\end{biography}

\end{document}